\documentclass[10pt,twocolumn,letterpaper]{article}

\usepackage[pagenumbers]{cvpr} 

\usepackage{graphicx}
\usepackage{amsmath}
\usepackage{amssymb}
\usepackage{booktabs}

\usepackage[pagebackref,breaklinks,colorlinks]{hyperref}

\usepackage[capitalize]{cleveref}
\crefname{section}{Sec.}{Secs.}
\Crefname{section}{Section}{Sections}
\Crefname{table}{Table}{Tables}
\crefname{table}{Tab.}{Tabs.}

\usepackage{graphicx}
\usepackage{booktabs}
\usepackage{algorithm}
\usepackage{algpseudocode}
\usepackage{diagbox}
\usepackage{xcolor}
\usepackage{comment}
\usepackage{enumitem}

\usepackage{amsmath}
\usepackage{amsfonts}

\usepackage[accsupp]{axessibility}  

\newcommand{\Hce}{\mathrm{H}^{\tiny{\text{xe}}}}
\newcommand{\xv}{\mathbf{x}}

\newcommand{\mv}{\mathbf{m}}
\newcommand\ourmodel{PETAL}

\def\cvprPaperID{10567} 
\def\confName{CVPR}
\def\confYear{2022}

\begin{document}

\title{A Probabilistic Framework for Lifelong Test-Time Adaptation}

\author{Dhanajit Brahma\\
Indian Institute of Technology Kanpur\\
{\tt\small dhanajit@cse.iitk.ac.in}
\and
Piyush Rai\\
Indian Institute of Technology Kanpur\\
{\tt\small piyush@cse.iitk.ac.in}
}
\maketitle

\begin{abstract}
Test-time adaptation (TTA) is the problem of updating a pre-trained source model at inference time given test input(s) from a different target domain. Most existing TTA approaches assume the setting in which the target domain is \emph{stationary}, i.e., all the test inputs come from a single target domain. However, in many practical settings, the test input distribution might exhibit a lifelong/continual shift over time. Moreover, existing TTA approaches also lack the ability to provide reliable uncertainty estimates, which is crucial when distribution shifts occur between the source and target domain. To address these issues, we present PETAL (Probabilistic lifElong Test-time Adaptation with seLf-training prior), which solves lifelong TTA using a probabilistic approach, and naturally results in (1) a student-teacher framework, where the teacher model is an exponential moving average of the student model, and (2) regularizing the model updates at inference time using the source model as a regularizer. To prevent model drift in the lifelong/continual TTA setting, we also propose a data-driven parameter restoration technique which contributes to reducing the error accumulation and maintaining the knowledge of recent domains by restoring only the irrelevant parameters.  In terms of predictive error rate as well as uncertainty based metrics such as Brier score and negative log-likelihood, our method achieves better results than the current state-of-the-art for online lifelong test-time adaptation across various benchmarks, such as CIFAR-10C, CIFAR-100C, ImageNetC, and ImageNet3DCC datasets. 
The source code for our approach is accessible at \url{https://github.com/dhanajitb/petal}.
\end{abstract}

\section{Introduction}
Deep learning models exhibit excellent performance in settings where the model is evaluated on data from the same distribution as the training data.
However, the performance of such models degrades drastically when the distribution of the test inputs at inference time is different from the distribution of the train data (source)~\cite{hendrycks2018benchmarking,taori2020measuring,koh2021wilds}.
Thus, there is a need to robustify the network to handle such scenarios. A particularly challenging setting is when we do not have any labeled target domain data to  finetune the source model, and unsupervised adaptation must happen at test time when the unlabeled test inputs arrive. This problem is known as \emph{test-time adaptation} (TTA)~\cite{wang2020tent,mummadi2021test,sun2020test}.
Moreover, due to the difficulty of training a single model to be robust to all potential distribution changes at test time, standard fine-tuning is infeasible, and TTA becomes necessary. Another challenge in TTA is that the source domain training data may no longer be available due to privacy/storage requirements, and we only have access to the source pre-trained model.

\begin{figure*}[!htbp]
    \centering
    \includegraphics[width=1.0\textwidth]{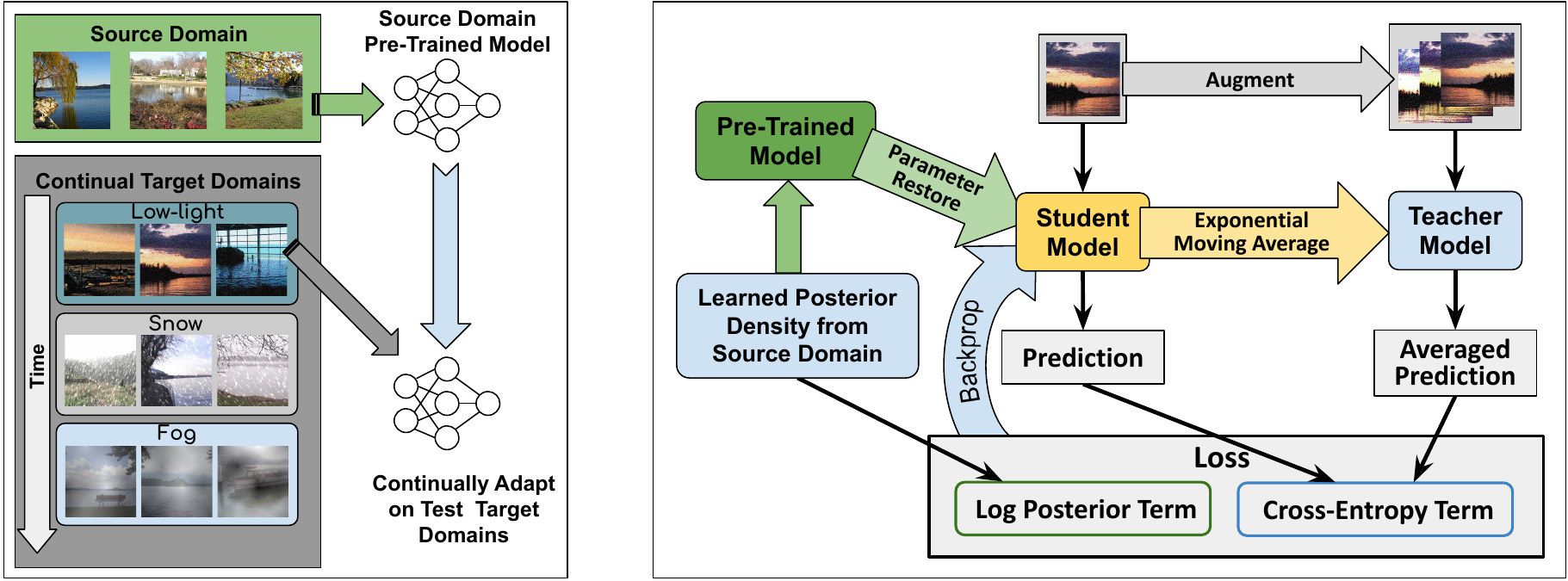}
    \caption{\textbf{Left:} Problem setup of online lifelong TTA.
    During adaptation on test input, the source domain data is no longer available, and only the model pre-trained on the source domain is provided. 
    Test inputs from different domains arrive continually, and the model has no knowledge about change in the domain.
    \textbf{Right:} Our proposed probabilistic framework for online lifelong TTA.
    We obtain a source domain pre-trained model from the posterior density learned using training data from the source domain.
    The posterior density is used to initialize the student model. 
    A test sample is provided as input to the student model.
    Using multiple augmentations of a test sample, we obtain augmentation averaged prediction from the teacher model.
    The loss term consists of log posterior and cross-entropy terms utilizing student and teacher model predictions.
    We utilize backpropagation to update student model and exponential moving average for teacher model.
    }
    \label{fig:taskframework}
\end{figure*}

Current approaches addressing the problem of TTA \cite{wang2020tent,mummadi2021test,sun2020test,wang2022continual} are based on techniques like self-training based pseudo-labeling or entropy minimization in order to enhance performance under distribution shift during testing.
One crucial challenge faced by existing TTA methods is that real-world machine learning systems work in non-stationary and continually changing environments.
Even though the self-training based approaches perform well when test inputs are from a different domain but all still i.i.d., it has been found that the performance is unstable when target test inputs come from a continually changing environment \cite{prabhu2021sentry}. Thus, it becomes necessary to perform test-time adaptation in a continual manner.

Such a setting is challenging because the continual adaptation of the model in the long term makes it more difficult to preserve knowledge about the source domain.
Continually changing test distribution causes pseudo-labels to become noisier and miscalibrated \cite{guo2017calibration} over time, leading to error accumulation \cite{chen2019progressive} which is more likely to occur if early predictions are incorrect.
When adapting to new test input, the model tends to forget source domain knowledge, triggering \textit{catastrophic forgetting} \cite{mccloskey1989catastrophic,ratcliff1990connectionist,parisi2019continual}.
Moreover, existing TTA methods do not account for \emph{model/predictive uncertainty}, which can result in miscalibrated predictions.

Recently, \cite{wang2022continual} proposed CoTTA, an approach to address the continual/lifelong TTA setting using a stochastic parameter reset mechanism  to prevent forgetting. Their reset mechanism however is based on randomly choosing a subset of weights to reset and is not data-driven. Moreover, their method does not take into account model/predictive uncertainty and is therefore susceptible to overconfident and miscalibrated predictions.

To improve upon these challenges of continual/lifelong TTA, we propose a  principled, probabilistic framework for lifelong TTA. Our framework (shown in Fig.~\ref{fig:taskframework} (Right)) constructs a posterior distribution over the source model weights and a data-dependent prior which results in a self-training based cross-entropy loss, with a regularizer term in the learning objective.
This regularizer arises from terms corresponding to the posterior, which incorporates knowledge of source (training) domain data.

Moreover, our framework also offers a probabilistic perspective and justification to the recently proposed CoTTA ~\cite{wang2022continual} approach, which arises as a special case of our probabilistic framework. In particular, only considering the data-driven prior in our approach without the regularizer term, corresponds to the student-teacher based cross-entropy loss used in CoTTA.
Further, to improve upon the stochastic restore used by \cite{wang2022continual}, we present a \emph{data-driven} parameter restoration based on Fisher Information Matrix (FIM).
In terms of improving accuracy and enhancing calibration during distribution shift, our approach surpasses existing approaches in various benchmarks.
\vspace{0.5em}\\
\textbf{Main Contributions}
\begin{enumerate}[topsep=3px,partopsep=0px,itemsep=0px]
    \item From a probabilistic perspective, we arrive at the student-teacher training framework in our proposed Probabilistic lifElong Test-time Adaptation with seLf-training prior (\ourmodel{}) approach. 
    Inspired from the self-training framework \cite{lee2013pseudo,xie2020self}, the teacher model is the exponential moving average of the student model, as depicted in Fig.~\ref{fig:taskframework} (Right).
    \item The student-teacher cross-entropy loss with a regularizer term corresponding to posterior of source domain data naturally emerges in the probabilistic formulation.
    \item We propose a data-driven parameter restoration based on Fisher Information Matrix (FIM) to handle error accumulation and catastrophic forgetting.
\end{enumerate}

\section{Problem Setup and Background}
In this section, we define the notation used, the problem setup of lifelong/continual TTA, and the basic probabilistic framework on which our approach is based. 

\subsection{Problem Setup}
Let $\mathbf{x}$ denote the inputs sampled i.i.d from a generative model having parameters $\psi$, and $p(y|\mathbf{x}, \theta)$, having parameters $\theta$, be the conditional distribution from which the corresponding labels are sampled. We denote the prior distributions of $\theta$ and $\psi$ using $p(\theta)$ and $p(\psi)$, respectively.

A typical test-time adaptation setting is as follows: We have a model with parameters $\theta_0$ trained on the source training data $\mathcal{X} = \{ \mathbf{x}_n,y_n\}_{n=1}^{N}$. The aim is to adapt $\theta_0$ and perform well on the test inputs $\mathcal{U}^d = \{ \mathbf{x}_m^d\}_{m=1}^{M_d}$ from an unlabeled target domain $d$. In case of multiple target domains, the adaptation happens for each target domain separately: 
$\theta_0 \rightarrow \theta_{d}$.

In \emph{lifelong/continual} test-time adaptation, unlabeled test inputs from $D$ different target domains $\{\mathcal{U}^d: d=1,\cdots,D\}$ arrive continually, and, thus, the model can utilize only the data available for the current target domain. Note that there is no information available to the learner about the change in domain.
At step $t$, for a test input $\mathbf{x}_m$, we make predictions using $p(y_m|\mathbf{x}_m,\theta_t)$ as well as adapt the parameters for future steps, i.e., $\theta_t \rightarrow \theta_{t+1}$.
Note that there is a continual domain shift in the data distribution of $\mathbf{x}_m$. Moreover, the model evaluation is performed based on predictions obtained online. Fig.~\ref{fig:taskframework} (Left) depicts the online lifelong TTA problem setup.

\subsection{The Underlying Probabilistic Framework}
In this section, we review standard probabilistic discriminative models for supervised learning and semi-supervised learning (SSL), which incorporates unlabeled data via a \emph{partly} data-dependent prior. Then we discuss a formulation of self-training based Bayesian SSL with partly data-driven cross-entropy prior. 
Further, we describe a modification to this Bayesian SSL formulation to handle the situation when the distribution of unlabeled inputs is different from the labeled inputs (covariate shift). 

In Section~\ref{sec:tta}, we describe how this Bayesian SSL formulation can be further extended for our problem, i.e., the lifelong/continual TTA setting where we learn a source model using only labeled data, and then the model has to predict labels of unlabeled test inputs coming from target domains with different distributions.
\vspace{0.5em}\\
\textbf{Bayesian Supervised Learning}\\
The Bayesian setup for supervised learning typically assumes that we are given labeled data $\mathcal{D} = \{ \mathbf{x}_n,y_n\}_{n=1}^N$, and we estimate $\theta$ using its posterior distribution
\begin{equation}
    p(\theta|\mathcal{D}) \propto p(\theta) \prod_{n=1}^{N} p(y_n|\mathbf{x}_n,\theta).
    \label{bayessup}
\end{equation}

Given a novel test input $\mathbf{x}_{N+1}$, we make predictions using posterior predictive distribution obtained by marginalizing over the posterior distribution
$
    p(y_{N+1}|\mathbf{x}_{N+1}) = \int p(y_{N+1}|\mathbf{x}_{N+1},\theta)p(\theta|\mathcal{D}) d\theta
$.
\vspace{0.5em}\\
\textbf{Bayesian Semi-Supervised Learning}\\
Here, we are provided with unlabeled data along with some labeled data. We denote the unlabeled data points as~~$\mathcal{U}=\{ \mathbf{x}_m\}_{m=1}^M$. To circumvent the inability to use unlabeled data while inferring $\theta$, one needs to make assumptions about the dependency between distributions of inputs and labels.

To this end, \cite{grandvalet2004semi} proposed a prior that is partly data-dependent via the inputs $\xv$: $p(\theta|\psi) \propto p(\theta) \exp(-\lambda \mathrm{H}_{\theta,\psi}(y|\xv)),$ where $p(\theta)$ is the prior in Eq.~\ref{bayessup}, and $\mathrm{H}$ is the conditional entropy of the class label. Here, \emph{partly data-dependent} prior refers to a prior defined using only the input $\xv$ treated as given, not the label $y$ which is treated as a random variable.
\vspace{0.5em}\\
\textbf{Bayesian Semi-Supervised Learning with Self-Training}\\
The self-training framework \cite{lee2013pseudo,xie2020self} has demonstrated significant success in semi-supervised learning. Our proposed framework is also based on self-training wherein we use an exponential moving average of the parameters $\theta$ of the student model $p(y|\xv,\theta$) (which is initialized with the source pre-trained model parameters $\theta_0$), and refer to the averaged model as the teacher model $(\theta')$:
$
    \theta'_{t+1} = \pi\theta'_{t} + (1-\pi)\theta_{t+1}
$, where $\pi$ is the smoothing factor.
For brevity, we will omit the time step index $t$ from here onwards.

In the semi-supervised setting, the teacher model can be utilized to obtain augmentation-averaged pseudo-labels $y'$ corresponding to an unlabeled input $\mathbf{x}$.
To prevent error accumulation, augmentation is only used when the domain difference is substantial. Defining
    $\hat{y}' = p(y|\xv,\theta')$; 
    $\tilde{y}' = \frac{1}{K}\sum_{i=1}^K p(y|\alpha_i(\xv),\theta')$, the pseudo-label is defined as 
\begin{align}
    y' = &\begin{cases}
            \hat{y}', & \text{if } C(p(y|\xv,\theta_0)\geq \tau)\\
            \tilde{y}', & \text{otherwise}.
          \end{cases}
\end{align}
Here, $K$ is the number of times augmentation is applied, $\alpha_i()$ is the augmentation function, $C()$ gives the confidence of the prediction, and $\tau$ is the threshold for selecting confident predictions.
The prediction confidence of current input using source domain pre-trained model $\theta_0$ gives us an estimate of the domain difference between the source and target domain.

Using these pseudo labels, we formulate the following partly data-driven cross-entropy prior
\begin{equation}
\begin{aligned}
    p(\theta|\psi) \propto~& p(\theta) \exp(-\lambda \Hce_{\theta,\psi}(y', y|\mathbf{x})) \\
    =~ p(\theta) \exp&(\lambda \mathbb{E}_{\xv\sim p(\xv|\psi),y'\sim p(y|\xv,\theta')}[\log p(y|\xv,\theta)]).
    \label{xeprior}
\end{aligned}
\end{equation}
Here, $y = p(y|\xv,\theta)$ is the prediction of the student model and $\Hce$ is the conditional cross-entropy of labels conditioned on the inputs. This cross-entropy term leverages the knowledge from the teacher model as it is incorporated into the partly data-driven prior.
\vspace{0.5em}\\
\textbf{Bayesian Semi-Supervised Learning with Unlabeled Data Distribution Shift}\\ 
The above semi-supervised learning formulation assumes that the unlabeled inputs come from the same distribution as the labeled inputs. To handle the situation when the unlabeled inputs come from a different distribution, we introduce additional generative parameters $\bar{\psi}$ while using the same conditional model $p(y|\xv,\theta)$ parameters $\theta$ for both distributions. Unlabeled inputs $\mathbf{\bar{x}}$ which come from a different distribution, are assumed to be sampled from the generative model with parameters $\bar{\psi}$.

Incorporating the additional generative parameters $\bar{\psi}$ in Eq.~\ref{xeprior}, the prior becomes
\begin{align}
    p(\theta|\psi,\bar{\psi}) \propto p(\theta) & \exp(-\lambda \Hce_{\theta,\psi}(y', y|\mathbf{x})) \nonumber \\
& \exp(-\bar{\lambda} \Hce_{\theta,\bar{\psi}}(y', y|\mathbf{\bar{x}})).
    \label{xeteprior}
\end{align}

The conditional entropies present in Eq.~\ref{xeteprior} require expectations over the distributions of the (labeled and unlabeled) inputs.

Replacing $p(\xv|\psi)$ and $p(\bar{\xv}|\bar{\psi})$ with the empirical distributions of $\xv$ and $\bar{\xv}$, we get
\begin{align}
    p(\theta|\psi,\bar{\psi}) & \propto p(\theta) \exp\left(-\frac{\lambda}{N} \sum_{n=1}^N\Hce(y'_n, y_n|\mathbf{x}_n,\theta)\right) \label{xeteempprior} \nonumber \\ 
& \exp\left(-\frac{\bar{\lambda}}{M} \sum_{m=1}^M\Hce(y'_m, y_m|\mathbf{\bar{x}}_m,\theta)\right).
\end{align}

Using Eq.~\ref{xeteempprior} as the prior in Eq.~\ref{bayessup} and taking logarithm on both sides (ignoring additive normalization constants), the new posterior distribution of the network parameters becomes: $\log p(\theta|\mathcal{D}, \mathcal{U}) = \log p(\theta) + \sum_{n=1}^N\log p(y_n|\xv_n,\theta) - \frac{\lambda}{N} \sum_{n=1}^N\Hce(y'_n, y_n|\mathbf{x}_n) - \frac{\bar{\lambda}}{M} \sum_{m=1}^M\Hce(y'_m, y_m|\mathbf{\bar{x}}_m)$. 

Since the labeled data is already used in the likelihood term, in the prior we ignore the cross-entropy term for the labeled data by setting $\lambda=0$. Thus, the log-posterior density is simplified to
\begin{align}
    \log p(\theta|\mathcal{D}, \mathcal{U}) = \log p(\theta) 
    + & \sum_{n=1}^N\log p(y_n|\xv_n,\theta) \nonumber \\
    -\frac{\bar{\lambda}}{M} & \sum_{m=1}^M\Hce(y'_m, y_m|\mathbf{\bar{x}}_m).
    \label{logsimprior}
\end{align}
\section{Probabilistic Test-Time Adaptation}
\label{sec:tta}
Our formulation for TTA is similar to the formulation we described above for the problem of Bayesian semi-supervised learning with unlabeled data distribution shift, with a key difference. In contrast to Bayesian SSL in which labeled and unlabeled data are available during training, in \textit{test-time adaptation}, we need to predict labels of the test inputs from a different domain, but we only have access to the source model weights without the source domain training data. In our probabilistic approach, we assume that the source model is given in form of the approximate posterior distribution $q(\theta)$ over the weights of the source model
\begin{equation}
    q(\theta) \approx p(\theta|\mathcal{D}).
    \label{apppost}
\end{equation}

At test-time, we use $q(\theta)$ to represent the source domain knowledge. 
Substituting Eq.~\ref{bayessup} for the posterior of the source domain data, we take the logarithm in Eq.~\ref{apppost} and simplify further to get the following
\begin{equation}
    \log q(\theta) = \log p(\theta) + \sum_{n=1}^N\log p(y_n|\xv_n,\theta).
\end{equation}

For our TTA setting, substituting this approximate posterior above in Eq.~\ref{logsimprior}, the log-posterior density with both labeled (source) and unlabeled (test inputs) data becomes
\begin{equation}
\log p(\theta|\mathcal{D}, \mathcal{U}) = \log q(\theta) -\frac{\bar{\lambda}}{M} \sum_{m=1}^M\Hce(y'_m, y_m|\mathbf{\bar{x}}).\label{finobj}
\end{equation}

Since posterior inference for deep neural networks is challenging, we leverage the Gaussian posterior approximation based on the SWAG-diagonal~\cite{maddox2019simple} method. It uses the SGD iterates to construct the mean and (diagonal) covariance matrix of the Gaussian posterior approximation and requires minimal changes to the training mechanism on source domain training data.
\subsection{Parameter Restoration}
\hspace{-1.2em}\textbf{Stochastic Restoration}\\
In lifelong TTA, in order to reduce the error accumulation over the long term in self-training and handle catastrophic forgetting, \cite{wang2022continual} proposed stochastic restoration of weights by additionally updating the parameters.
Let $\theta^f$ denote the flattened parameter $\theta$ of the student model, and $D$ be the dimension of $\theta^f$.
After the gradient update at time step $t$, stochastic restore further updates the parameters:
\begin{align}
    &\mv \sim \text{Bernoulli}(\rho), \label{sresbern} \\
    \theta^f_{t+1}& = \mv \odot \theta^f_0 + (\mathbf{1}-\mv) \odot \theta^f_{t+1}.
\end{align}
Here, $\odot$ is element-wise multiplication, and $\rho$ is stochastic restore probability. $\mv$ is mask to determine which parameters within $\theta^f_{t+1}$ to restore to original source weight $\theta^f_0$.
\vspace{0.5em}\\
\textbf{Fisher Information Based Restoration}\\
To improve upon stochastic restoration, we propose a data-driven parameter restoration.
Fisher Information Matrix (FIM) is widely used as a metric of parameter importance for a given data \cite{kirkpatrick2017overcoming}.
Thus, we use FIM, $F$, of the student model parameterized by $\theta$ as a measure of the importance of the parameters.
\begin{algorithm}[!htbp]
    \caption{Proposed Approach \ourmodel{}}
    \label{alg:bcotta}
    \begin{algorithmic}[1]
        \Statex \hspace{-1.68em}\textbf{Input:} Training dataset $\mathcal{X} = \{ \mathbf{x}_n,y_n\}_{n=1}^N$;
        \Statex \hspace{1em} Test domain data $\mathcal{U}^d = \{ \mathbf{x}_m^d\}_{m=1}^{M_d}$, $d=1,\cdots,D$
        \Statex \hspace{-1.68em}\textbf{Output:} Prediction $\mathcal{Y}^d$
        \Statex \hspace{-1.68em}\textbf{Training:}
        \State \hspace{-1.68em}\centerline{Compute $q(\theta) \approx p(\theta|\mathcal{X})$}
        \Statex \hspace{-1.68em}\textbf{Continual Adaptation:}
        \State Initialize time step $t=0$, prediction set $\mathcal{Y}^d=\emptyset$
        \State Initialize student model $\theta_0 = \max_\theta \log q(\theta)$
        \State Initialize teacher model $\theta'_0=\theta_0$
        \For{$d=1,\cdots,D$}
        \For{\textbf{each batch} $\mathcal{B} \in \mathcal{U}^d$}
        \State $\theta'_{t+1},\theta_{t+1}, \mathcal{B}_y$ = Adapt($\theta'_t,\theta_t,\theta_0,\mathcal{B},q(\theta),t$)
        \State $\mathcal{Y}^d = \mathcal{Y}^d \cup \mathcal{B}_y$
        \State $t=t+1$
        \EndFor
        \EndFor
    \end{algorithmic}
\end{algorithm}
\begin{algorithm}[!htbp]
    \caption{Adapt}
    \label{alg:augadapt}
    \begin{algorithmic}[1]
        \Statex \hspace{-1.68em}\textbf{Input:} $\theta'_t,\theta_t,\theta_0,\mathcal{B}, q(\theta),t$
        \Require Number of augmentations $K$; learning rate $\eta$;
        \Statex \hspace{2.3em} Threshold for confident predictions $\tau$;
        \Statex \hspace{2.3em} Confidence function $C$; Cross-entropy weight $\bar{\lambda}$;
        \Statex \hspace{2.3em} Quantile of FIM $\delta$; Augmentation function $\alpha$;
        \Statex \hspace{2.3em} Smoothing factor $\pi$
        \Statex \hspace{-1.68em}\textbf{Output:} $\theta'_{t+1},\theta_{t+1}, \mathcal{B}_y$
        \State Initialize: $\mathcal{H}=0$, $\mathcal{B}_y=\emptyset$
        \For{\textbf{each test input} $\mathbf{\bar{x}} \in \mathcal{B}$}
        \State Teacher model prediction: $\hat{y}' = p(y|\mathbf{\bar{x}},\theta')$
        \State Augmented Average: $\tilde{y}' = \frac{1}{K}\sum_{i=1}^K p(y|\alpha_i(\mathbf{\bar{x}}),\theta')$
        \State Augment based on domain gap:
        \begin{align}
        y' = &\begin{cases}
            \hat{y}', & \text{if } C(p(y|\mathbf{\bar{x}},\theta_0)\geq \tau)\\
            \tilde{y}', & \text{otherwise}.\nonumber
          \end{cases}
        \end{align}
        \vspace{-1.5em}
        \State Using student model, predict: $y = p(y|\xv,\theta)$
        \State Update: $\mathcal{H} = \mathcal{H} + \Hce(y', y|\mathbf{\bar{x}})$
        \State $\mathcal{B}_y = \mathcal{B}_y \cup \{y'\}$
        \EndFor
        
        \State Compute: $\mathcal{L} = \left(\log q(\theta) - \frac{\bar{\lambda}}{|\mathcal{B}|}\mathcal{H}\right)$
        \State Adapt student model: $\theta_{t+1}=\theta_t+\eta \nabla_\theta \mathcal{L}$ 
        \State Update teacher model: $\theta'_{t+1} = \pi\theta'_{t} + (1-\pi)\theta_{t+1}$
        \State Compute FIM:
        $
            F = \text{Diag}\left((\nabla_\theta\mathcal{L}) (\nabla_\theta\mathcal{L})^{\text{T}} \right)
        $
        \State Compute mask $\mv$ for resetting:
        \vspace{-0.75em}
        \begin{align}
                &\gamma = \text{quantile}(F, \delta) \nonumber \\
                \mv_p = &\begin{cases}
                        1, & \text{if $F_p<\gamma$}\\
                        0, & \text{otherwise}.
                      \end{cases}
                      ,~~p=1,\cdots,P. \nonumber
            \end{align}
        \State Reset updated student model back to source model:
        \vspace{-1.75em}
        \begin{align}
            \theta_{t+1} = \mv \odot \theta_0 + (\mathbf{1}-\mv) \odot \theta_{t+1} \label{sres} \nonumber
        \end{align}
    \end{algorithmic}
\end{algorithm}
For a given time step $t$ with unlabeled test inputs batch $\mathcal{B}=\{ \mathbf{\bar{x}}_m\}_{m=t|\mathcal{B}|}^{(t+1)|\mathcal{B}|}$, we consider the following diagonal approximation of FIM:
$
    F = \text{Diag}\left((\nabla_\theta\mathcal{L}) (\nabla_\theta\mathcal{L})^{\text{T}} \right)
$,
where,
    $
    \mathcal{L} = \log q(\theta) - \frac{\bar{\lambda}}{|\mathcal{B}|}\sum_{m=t|\mathcal{B}|}^{(t+1)|\mathcal{B}|}\Hce(y'_m, y_m|\mathbf{\bar{x}}_m)
    $.
Here, $y'_m$ and $y_m$ are the teacher and student model predictions, respectively. Note that $F$ has the same dimension as $\theta$.
Thus, upon using FIM based restore, the parameter restoration in Eq.~\ref{sresbern} becomes
\begin{align}
    \mv_p = &\begin{cases}
            1, & \text{if $F_p<\gamma$}\\
            0, & \text{otherwise}.
          \end{cases}
          , p=1,\cdots,P.
\end{align}
Here, $\gamma = \text{quantile}(F, \delta)$ is the threshold value which is the $\delta$-quantile of $F$.
Thus, the elements in $\mv$ corresponding to FIM value less than $\gamma$ would be 1, implying that the corresponding parameters would be restored to original source weight $\theta_f^{(0)}$.
The algorithm for \ourmodel{} is given in Algorithm~\ref{alg:bcotta}.
Fig.~\ref{fig:taskframework} (Right) provides an overview of our approach \ourmodel{}.

\section{Related Work}
\subsection{Unsupervised Domain Adaptation}
The goal of unsupervised domain adaptation (UDA) \cite{pan2010domain,ganin2015unsupervised,long2015learning,wang2018deep} is to enhance the performance of the learning model when there is a change in distribution between the training data domain and the test data domain.
UDA approaches often assume that the source and (unlabeled) target domain data are accessible simultaneously. Most existing methods address UDA by ensuring that the feature distributions \cite{long2015learning,ganin2015unsupervised,tsai2018learning} or the input spaces \cite{hoffman2018cycada,yang2020fda} of the source and the target domains are brought closer. 
\begin{table*}[!ht]
\centering
\scalebox{0.7}{
\tabcolsep7.2pt
\begin{tabular}{l|ccccccccccccccc|c}
\hline
Time & \multicolumn{15}{l|}{$t\xrightarrow{\hspace*{18.5cm}}$}& \\ \hline
Method & \rotatebox[origin=c]{70}{Gaussian} & \rotatebox[origin=c]{70}{shot} & \rotatebox[origin=c]{70}{impulse} & \rotatebox[origin=c]{70}{defocus} & \rotatebox[origin=c]{70}{glass} & \rotatebox[origin=c]{70}{motion} & \rotatebox[origin=c]{70}{zoom} & \rotatebox[origin=c]{70}{snow} & \rotatebox[origin=c]{70}{frost} & \rotatebox[origin=c]{70}{fog}  & \rotatebox[origin=c]{70}{brightness} & \rotatebox[origin=c]{70}{contrast} & \rotatebox[origin=c]{70}{elastic} & \rotatebox[origin=c]{70}{pixelate} & \rotatebox[origin=c]{70}{jpeg} & Mean \\ \hline
Source & 72.33 & 65.71 & 72.92 & 46.94 & 54.32 & 34.75 & 42.02 & 25.07 & 41.30 & 26.01 & 9.30 & 46.69 & 26.59 & 58.45 & 30.30 & 43.51 \\
BN Adapt   & 28.08 & 26.12 & 36.27 & 12.82 & 35.28 & 14.17 & 12.13 & 17.28 & 17.39 & 15.26 & 8.39 & 12.63 & 23.76 & 19.66 & 27.30 & 20.44 \\
Pseudo-label  &26.70&	22.10&	32.00&	13.80&	32.20&	15.30&	12.70&	17.30&	17.30&	16.50&	10.10&	13.40&	22.40&	18.90&	25.90&	19.80 \\
TENT-online$^+$ & 24.80 & 23.52 & 33.04 & 11.93 & 31.83 & 13.71 & 10.77 & 15.90 & 16.19 & 13.67 & 7.86 & 12.05 & 21.98 & 17.29 & 24.18 & 18.58 \\
TENT-continual & 24.80 &	\textbf{20.60} &	28.60 &	14.40 &	31.10 &	16.50 &	14.10 &	19.10 &	18.60 &	18.60 &	12.20 &	20.30 &	25.70 &	20.80 &	24.90 &	20.70\\
CoTTA & 23.92 & 21.40 & 25.95 & 11.82 & 27.28 & 12.56 & 10.48 & 15.31 & 14.24 & 13.16 & 7.69 & 11.00 & 18.58 & 13.83 & 17.17 & 16.29 (0.02)\\
\hline
\ourmodel{} (S-Res)  & 23.44 & 21.20 & \textbf{25.50} & 11.80 & \textbf{27.22} & 12.54 & 10.45 & 15.14 & 14.31 & 12.89 & 7.61 & 10.72 & 18.42 & 13.83 & 17.37 & 16.16 (0.02)\\
\ourmodel{} (FIM) & \textbf{23.42} & 21.13 & 25.68 & \textbf{11.71} & 27.24 & \textbf{12.19} & \textbf{10.34} & \textbf{14.76} & \textbf{13.91} & \textbf{12.65} & \textbf{7.39} & \textbf{10.49} & \textbf{18.09} & \textbf{13.36} & \textbf{16.81} & \textbf{15.95} (0.04)\\\hline
\end{tabular}
}
\caption{Experimental results for CIFAR10-to-CIFAR10C online lifelong test-time adaptation task. The numbers denote the classification error~(\%) obtained with the highest corruption of severity level 5. TENT-online uses domain information, denoted using +.} \label{tab:cifar10}
\end{table*}

\subsection{Test-Time Adaptation}
Some works also refer to TTA as \textit{source-free} domain adaptation.
Recent works explore source-free domain \cite{liang2020we,kundu2020universal,li2020model} setting in which training data is unavailable and only unlabeled data is available during adaptation. Test entropy minimization (TENT) \cite{wang2020tent} starts from a source pre-trained model and updates only the batch-norm (BN) parameters by minimizing entropy in test predictions. \cite{schneider2020improving} address TTA by updating the source domain BN statistics using test input statistics.
Continual Test-Time Adaptation (CoTTA)~\cite{wang2022continual} addresses online lifelong TTA by employing weight averaging and augmentation averaging, and random parameter restoration back to source pre-trained model parameters. 
\cite{gong2022robust} adapts to continually changing target domains by utilizing a normalization layer to handle the out-of-distribution examples and balanced reservoir sampling to store the simulated i.i.d. data in the memory. 
It would be an interesting future work to extend \ourmodel{} for the temporally correlated test stream setting proposed by \cite{gong2022robust}.
EATA~\cite{niu2022efficient} is another related work that looks at preventing forgetting in the context of TTA; however, EATA mainly focuses on preventing the forgetting of the source task model and is not designed to handle forgetting in a lifelong TTA setting.
Our work PETAL is a principled probabilistic approach for lifelong test-time adaptation that uses an approximate posterior during test-time adaptation obtained from source domain data. PETAL also offers a probabilistic perspective and justification to CoTTA which arises a special case of PETAL. 

Bayesian Adaptation for Covariate Shift (BACS) \cite{zhou2021training} proposes a Bayesian perspective for TENT, which naturally gives rise to the entropy term along with a regularizer that captures knowledge from posterior density obtained from training data. However, BACS only addresses standard TTA setting, and regularized entropy minimization lacks the ability to handle error accumulation and catastrophic forgetting encountered in lifelong TTA.

\subsection{Continual Learning}
The objective of Continual Learning (CL) is to learn from a sequential series of tasks, enabling the model to retain previously acquired knowledge while learning a new task, preventing catastrophic forgetting~\cite{mccloskey1989catastrophic,ratcliff1990connectionist,parisi2019continual}.
Elastic weight consolidation (EWC)~\cite{kirkpatrick2017overcoming} is a regularization-based technique that penalizes parameter changes having a significant impact on prediction.
\cite{li2017learning} proposed learning without forgetting (LwF), which preserves knowledge of previous tasks using knowledge distillation. 
Gradient episodic memory (GEM)~\cite{lopez2017gradient} maintains a limited number of samples to retrain while constraining fresh task updates from interfering with prior task knowledge.
\cite{brahma2021hypernetworks,wang2021ordisco} address the continual semi-supervised learning problem where continually arriving tasks consist of labeled and unlabeled data.

\begin{table*}[!ht]
\centering
\scalebox{0.7}{
\tabcolsep7.2pt
\begin{tabular}{l|lllllllllllllll|c}\hline
Time & \multicolumn{15}{l|}{$t\xrightarrow{\hspace*{18.5cm}}$}& \\ \hline
Method                          & \rotatebox[origin=c]{70}{Gaussian} & \rotatebox[origin=c]{70}{shot} & \rotatebox[origin=c]{70}{impulse} & \rotatebox[origin=c]{70}{defocus} & \rotatebox[origin=c]{70}{glass} & \rotatebox[origin=c]{70}{motion} & \rotatebox[origin=c]{70}{zoom} & \rotatebox[origin=c]{70}{snow} & \rotatebox[origin=c]{70}{frost} & \rotatebox[origin=c]{70}{fog}  & \rotatebox[origin=c]{70}{brightness} & \rotatebox[origin=c]{70}{contrast} & \rotatebox[origin=c]{70}{elastic} & \rotatebox[origin=c]{70}{pixelate} & \rotatebox[origin=c]{70}{jpeg} & Mean \\ \hline
Source & 73.00 & 68.01 & 39.37 & 29.32 & 54.11 & 30.81 & 28.76 & 39.49 & 45.81 & 50.30 & 29.53 & 55.10 & 37.23 & 74.69 & 41.25 & 46.45\\
BN Adapt & 42.14 & 40.66 & 42.73 & 27.64 & 41.82 & 29.72 & 27.87 & 34.88 & 35.03 & 41.50 & 26.52 & 30.31 & 35.66 & 32.94 & 41.16 & 35.37 \\
Pseudo-label &38.10 &	36.10&	40.70&	33.20&	45.90&	38.30&	36.40&	44.00&	45.60&	52.80&	45.20&	53.50&	60.10&	58.10&	64.50&	46.20\\
TENT-continual & \textbf{37.20} & \textbf{35.80} & 41.70 & 37.90 & 51.20 & 48.30 & 48.50 & 58.40 & 63.70 & 71.10 & 70.40 & 82.30 & 88.00 & 88.50 & 90.40 & 60.90 \\
CoTTA & 40.09 & 37.67 & 39.77 & 26.91 & 37.82 & 28.04 & 26.26 & 32.93 & 31.72 & 40.48 & 24.72 & 26.98 & 32.33 & 28.08 & 33.46 & 32.48 (0.02) \\
\hline
\ourmodel{} (S-Res)  & 38.37 & 36.43 & 38.69 & \textbf{25.87} & 37.06 & 27.34 & 25.55 & 32.10 & 31.02 & 38.89 & 24.38 & \textbf{26.38} & 31.79 & 27.38 & 32.98 & 31.62 (0.04)\\
\ourmodel{} (FIM) & 38.26 & 36.39  & \textbf{38.59} & 25.88 & \textbf{36.75} & \textbf{27.25} & \textbf{25.40} & \textbf{32.02} & \textbf{30.83} & \textbf{38.73} & \textbf{24.37} & 26.42 & \textbf{31.51} & \textbf{26.93} & \textbf{32.54} & \textbf{31.46} (0.04)\\\hline
\end{tabular}
}

\caption{Experimental results for CIFAR100-to-CIFAR100C online lifelong test-time adaptation task. The numbers denote the classification error rate~(\%) obtained with the highest corruption of severity level 5. }\label{tab:cifar100}
\end{table*}

\section{Experiments}
We thoroughly evaluate PETAL and compare it to other state-of-the-art approaches on image classification lifelong test-time adaptation benchmark tasks: CIFAR10-to-CIFAR10C, CIFAR100-to-CIFAR100C, ImageNet-to-ImageNetC, and ImageNet-to-Imagenet3DCC.

\subsection{Benchmark Datasets}
\cite{hendrycks2018benchmarking} developed CIFAR10C, CIFAR100C, and ImagenetC datasets to serve as benchmarks for the robustness of classification models.
In each dataset, there are 15 different types of corruption and five different levels of severity.
These corruptions are applied to test images of original CIFAR10 and CIFAR100 \cite{Krizhevsky09learningmultiple} datasets and validation images of original ImageNet \cite{deng2009imagenet} dataset.
Further, we experiment with Imagenet 3D Common Corruptions (Imagenet3DCC) dataset, recently proposed by \cite{kar20223d}, which utilizes the geometry of the scene in transformations, leading to more realistic corruptions. Imagenet3DCC dataset consists of 12 different types of corruptions, each with five levels of severity. Refer to Appendix for details.

In online lifelong TTA, we begin with a network trained on CIFAR10, CIFAR100, and ImageNet clean training set for the respective experiments.
At the time of testing, the model gets corrupted images online.

Following CoTTA, we continually adjust the source pre-trained model to each corruption type as they sequentially arrive, as opposed to conventional TTA in which the pre-trained model is separately adapted to each corruption type.
We evaluate the model using online predictions obtained immediately as the data is encountered.
We follow the online lifelong test-time adaptation setting for all the experiments. For ImageNet-to-ImageNetC experiments, we evaluate using 10 different sequences of corruptions.

\subsection{Model Architectures and Hyperparameters}
Following TENT \cite{wang2020tent} and CoTTA \cite{wang2022continual}, we adopt pre-trained WideResNet-28 \cite{zagoruyko2016wide} model for CIFAR10-to-CIFAR10C, pre-trained ResNeXt-29 \cite{xie2017aggregated} model for CIFAR100-to-CIFAR100C, and standard pre-trained ResNet-50 model for both ImageNet-to-ImagenetC and ImageNet-to-Imagenet3DCC experiments from RobustBench \cite{croce2021robustbench}.

We utilize SWAG-D~\cite{maddox2019simple} to approximate the posterior density from the source domain training data. SWAG-D approximates the posterior using a Gaussian distribution with diagonal covariance from the SGD trajectory.
Before adapting the model, we initialize it with the maximum a posteriori (MAP) of the approximate posterior that corresponds to the solution obtained by Stochastic Weight Averaging~\cite{izmailov2018averaging}. This is effectively the source domain pre-trained model.

We update all the trainable parameters in all experiments.
We use $K$ = 32 number of augmentations. We adopt the same augmentation confidence threshold described in \cite{wang2022continual}.
For FIM based parameter restoration, we set the quantile value $\delta$ = 0.03. We refer the readers to the Appendix for more details on the hyperparameters.
\subsection{Baselines and Compared Approaches}
To evaluate the efficacy of \ourmodel{}, we compare the PETAL with CoTTA and five other methods in online lifelong test-time adaptation.
\textit{Source} denotes the baseline pre-trained model that has not been adapted to test inputs.
In \textit{BN Adapt} \cite{li2016revisiting,schneider2020improving}, the network parameters are kept frozen, and only Batch Normalization statistics are adapted to produce predictions for test inputs.
\textit{Pseudo-label} updates the BatchNorm trainable parameters with hard pseudo-labels \cite{lee2013pseudo}.
\textit{TENT-online} denotes the TENT \cite{wang2020tent} approach in this setting, but it has access to extra information about the change in the domain and, thus, resets itself to the original pre-trained model upon encountering test inputs from the new domain and adapts afresh.
But this additional knowledge is unavailable in real-life scenarios.
\textit{TENT-continual} has no extra information about domain change.
\textit{CoTTA} \cite{wang2022continual} uses weight averaging and augmentation averaging, along with randomly restoring parameters to the original pre-trained model.
However, it lacks explicit uncertainty modeling and data-driven parameter restoration.
\subsection{Evaluation Metrics}
We evaluate our model using the error rate in predictions.
To evaluate the uncertainty estimation, we use negative log-likelihood (NLL) and Brier score~\cite{brier1950verification}.
Both NLL and Brier are proper scoring rules \cite{gneiting2007strictly}, and they are minimized if and only if the predicted distribution becomes identical to the actual distribution.
In Table~\ref{tab:cifar10},~\ref{tab:cifar100} and~\ref{tab:unccifar100}, the number within brackets is the standard deviation over 5 runs.
Refer to the Appendix for more details on evaluation metrics.
\begin{table}[!htb]
\caption{CIFAR10-to-CIFAR10C results for gradually changing severity level before changing corruption types. The numbers are averaged over all 15 corruption types. The number after $\pm$ is the standard deviation over 10 random corruption sequences. Our method surpasses all baselines in the depicted settings.}\label{tab:gradualcifar10}
\small
\centering
\scalebox{0.7}{
\tabcolsep 6pt
\begin{tabular}{c|c|c|c|c|c}
\hline
\diagbox{Metric}{Method} & Source & BN Adapt & TENT & CoTTA & \ourmodel{} (FIM) \\
\hline
Error (\%) & 23.94 & 13.54 & 29.46 & 10.40 $\pm$ 0.22 & \textbf{10.11} $\pm$ \textbf{0.23} \\
Brier & 0.408 & 0.222 & 0.575 & 0.159 $\pm$ 0.003 & \textbf{0.158} $\pm$ \textbf{0.004} \\
\hline
\end{tabular}}
\end{table}
\subsection{CIFAR10-to-CIFAR10C Results}
In Table~\ref{tab:cifar10}, we observe that directly using the pre-trained model (\textit{Source}) leads to poor performance with an average error rate of 43.51\%, suggesting the necessity of adaptation.
Adapting the Batch Normalization (BN) statistics improves the average error rate to 20.44\%.
Using hard pseudo-labels and updating only the BN parameters further improves the performance to 19.8\%.
TENT-online reduces the error rate to 18.58\% using extra information about domain change, but access to such information is mostly unavailable in real-world scenarios.
As expected, the error rate of TENT-continual increases to 20.7\% without access to domain change information.
Further, CoTTA improves the average error rate to 16.29\%.
Our proposed approach consistently outperforms other approaches for most individual corruption types, reducing the average error rate to 15.95\%. Moreover, our proposed approach demonstrates no performance degradation in the long term. 

To investigate the contribution of FIM based parameter restore, we show the results of our approach with stochastic restore in place of FIM based restore, denoted using S-Res. We observe that FIM based restore performs better than S-Res for most of the corruption types, highlighting the effectiveness of FIM based restore.
Further, the improved performance of \ourmodel{} (S-Res) and \ourmodel{} (FIM) over various baselines, including CoTTA which is specifically designed for continual TTA, demonstrates the effectiveness of our probabilistic framework where the source model's posterior induces a regularizer and the data-driven resetting helps make an informed selection of weights to reset/keep.
Moreover, the contribution of the regularizer term (induced by the source model’s posterior) is evident from the superior performance of PETAL over CoTTA, since CoTTA is a special case of PETAL without the regularizer term.
\vspace{0.5em}\\
\textbf{Gradually changing corruptions:} 
Following~\cite{wang2022continual}, we evaluate \ourmodel{} in the setting where the severity of corruption changes gradually before the change in corruption type.
When the corruption type changes, the severity level is lowest, and thus, domain shift is gradual. In addition, distribution shifts within each corruption type are also gradual. Refer to the Appendix for details.

We report the average error and average Brier score over 10 randomly shuffled orders of corruptions. We can observe in Table~\ref{tab:gradualcifar10} that our approach \ourmodel{} performs better than other approaches in terms of error and Brier score.

\begin{table}[!htb]
\caption{CIFAR100-to-CIFAR100C results with the most severe level of corruption, 5, averaged over all corruption types. Our method surpasses all baselines in terms of NLL and Brier uncertainty estimation measures.}\label{tab:unccifar100}
\small
\centering
\scalebox{0.7}{
\tabcolsep 5pt
\begin{tabular}{c|c|c|c|c|c}
\hline
\diagbox{Metric}{Method} & Source & BN Adapt & TENT & CoTTA & \ourmodel{} (FIM) \\
\hline
NLL & 2.4945 & 1.3937 & 7.3789 & 1.2767 (0.0006) & \textbf{1.2206} (0.0011) \\
Brier & 0.6704 & 0.4744 & 1.1015 & 0.4430 (0.0003) & \textbf{0.4317} (0.0003) \\
\hline
\end{tabular}}
\end{table}
\subsection{CIFAR100-to-CIFAR100C Results}
To illustrate the efficacy of the proposed approach, we conduct an evaluation on the more challenging CIFAR100-to-CIFAR100C task. In Table~\ref{tab:cifar100}, we compare the results with Source, BN Adapt, Pseudo-label, and TENT-continual approaches. We observe that TENT-continual performs better initially, but as new domains arrive continually, the performance degrades drastically in the long term. With an average error rate of 31.46\%, our proposed approach \ourmodel{} (FIM) consistently outperforms the other approaches.

To measure the ability of uncertainty estimation of our approach, we compare with the other approaches in Table~\ref{tab:unccifar100} in terms of NLL and Brier score. We obtain both NLL and Brier score for corruption with a severity level of 5 and average over all corruption types. Our approach performs better than all other approaches in terms of average NLL and average Brier, demonstrating the ability of our approach to improve the uncertainty estimation. 
Moreover, the FIM based restore outperforms stochastic restore for most corruption types, illustrating the utility of data-driven resetting.
\begin{figure}[!htb]
  \centering
  \includegraphics[width=0.95\linewidth]{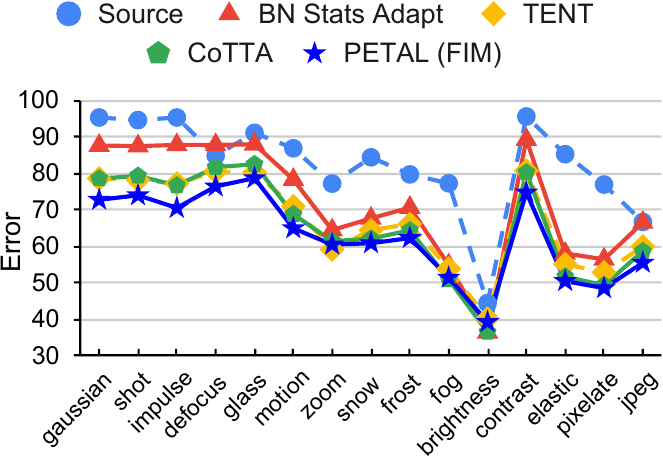}\\
  \vspace{0.1em}
  \includegraphics[width=0.95\linewidth]{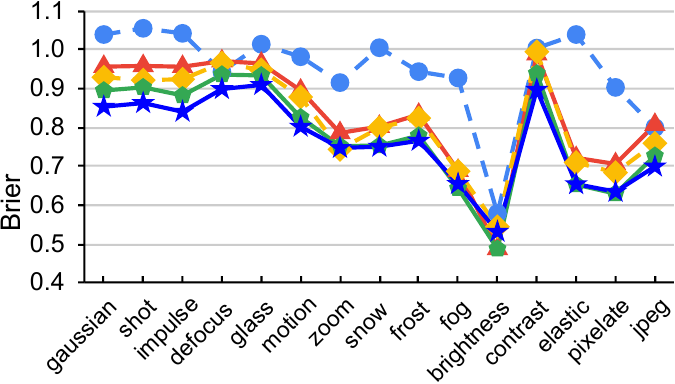}
\caption{ImageNet-to-ImageNetC results averaged over 10 different corruption orders with level 5 corruption severity.}
\label{fig:imagenetres}
\end{figure}
\subsection{ImageNet-to-ImageNetC Results}
In Table~\ref{tab:imagenet}, we investigate the performance of our proposed approach for ImageNet-to-ImageNetC task with 10 different sequences of corruption. We obtain the performance scores by averaging over all corruption types and all corruption orders. In terms of average error, average NLL, and average Brier, our approach performs better than the other approaches. In Fig.~\ref{fig:imagenetres}, we compare the performance by averaging over 10 different corruption orders in terms of error rate and Brier score. For most corruption types, our approach performs better than other existing approaches. 

\begin{table}[!htb]
\caption{ImageNet-to-ImageNetC results averaged over all corruption types and over 10 diverse corruption orders (highest corruption severity level 5).}\label{tab:imagenet}
\small
\centering
\scalebox{0.7}{
\tabcolsep 7.5pt
\begin{tabular}{c|c|c|c|c|c}
\hline
\diagbox{Metric}{Method} & Source & BN Adapt & TENT & CoTTA & \ourmodel{} (FIM) \\
\hline
Error (\%) & 82.35 & 72.07 & 66.52 & 63.18 & \textbf{62.71} \\
NLL & 5.0701 & 3.9956 & 3.6076 & 3.3425 & \textbf{3.3252} \\
Brier & 0.9459 & 0.8345 & 0.8205 & 0.7681 & \textbf{0.7663} \\
\hline
\end{tabular}}
\end{table}
\subsection{ImageNet-to-ImageNet3DCC Results}
For ImageNet-to-ImageNet3DCC dataset, we experiment with 10 different random sequences of corruptions. We provide the results in Table~\ref{tab:imagenet3d} by averaging over 10 random sequences of corruptions and 12 corruption types at severity level 5. \ourmodel{} consistently outperforms all other approaches in terms of error rate, NLL, and Brier score.
\begin{table}[!htb]
\caption{ImageNet-to-ImageNet3DCC results averaged over all corruption types and over 10 diverse corruption orders (highest corruption severity level 5).}\label{tab:imagenet3d}
\small
\centering
\scalebox{0.7}{
\tabcolsep 7.5pt
\begin{tabular}{c|c|c|c|c|c}
\hline
\diagbox{Metric}{Method} & Source & BN Adapt & TENT & CoTTA & \ourmodel{} (FIM) \\
\hline
Error (\%) & 69.21 & 67.32 & 95.93 & 59.91 & \textbf{59.61} \\
NLL & 5.0701 & 3.9956 & 3.6076 & 3.3425 & \textbf{3.3252} \\
Brier & 3.9664 & 3.7163 & 19.0408 & 3.2636 & \textbf{3.2560} \\
\hline
\end{tabular}}
\end{table}
\section{Conclusion}
We proposed a probabilistic framework for lifelong TTA using a partly data-driven prior.
Addressing the problem via the probabilistic perspective naturally gives rise to the student-teacher framework along with a regularizer that captures the source domain knowledge.
In lifelong TTA, we have demonstrated that our principled use of an approximate training posterior surpasses prior heuristic approaches.
Our proposed approach also provides more reliable uncertainty estimates demonstrated with better NLL and Brier score.
Further, we developed a Fisher information matrix based parameter restoration, which is driven by the data to improve upon existing stochastic restore.
In terms of error rate, NLL and Brier score, PETAL yields state-of-the-art results across CIFAR10-to-CIFAR10C, CIFAR100-to-CIFAR100C, ImageNet-to-ImageNetC, and ImageNet-to-ImageNet3DCC benchmark tasks.
\vspace{0.5em}\\
\textbf{Acknowledgments}\\
The authors acknowledge support from the Qualcomm Innovation Fellowship and travel grant support from the Research-I Foundation at IIT Kanpur.

\bibliographystyle{ieee_fullname}
\bibliography{references}

\begin{thebibliography}{10}\itemsep=-1pt

\bibitem{brahma2021hypernetworks}
Dhanajit Brahma, Vinay~Kumar Verma, and Piyush Rai.
\newblock Hypernetworks for continual semi-supervised learning.
\newblock In {\em International Workshop on Continual Semi-Supervised Learning,
  IJCAI}, 2021.

\bibitem{brier1950verification}
Glenn~W. Brier.
\newblock {Verification of forecasts expressed in terms of probability}.
\newblock {\em Monthly Weather Review}, 78(1), 1 1950.

\bibitem{chen2019progressive}
Chaoqi Chen, Weiping Xie, Wenbing Huang, Yu Rong, Xinghao Ding, Yue Huang,
  Tingyang Xu, and Junzhou Huang.
\newblock Progressive feature alignment for unsupervised domain adaptation.
\newblock In {\em Proceedings of the IEEE/CVF Conference on Computer Vision and
  Pattern Recognition}, pages 627--636, 2019.

\bibitem{cohen2021katana}
Gilad Cohen and Raja Giryes.
\newblock Katana: Simple post-training robustness using test time
  augmentations.
\newblock {\em arXiv preprint arXiv:2109.08191}, 2021.

\bibitem{croce2021robustbench}
Francesco Croce, Maksym Andriushchenko, Vikash Sehwag, Edoardo Debenedetti,
  Nicolas Flammarion, Mung Chiang, Prateek Mittal, and Matthias Hein.
\newblock Robustbench: a standardized adversarial robustness benchmark.
\newblock In {\em Thirty-fifth Conference on Neural Information Processing
  Systems Datasets and Benchmarks Track (Round 2)}, 2021.

\bibitem{deng2009imagenet}
Jia Deng, Wei Dong, Richard Socher, Li-Jia Li, Kai Li, and Li Fei-Fei.
\newblock Imagenet: A large-scale hierarchical image database.
\newblock In {\em 2009 IEEE Conference on Computer Vision and Pattern
  Recognition}, pages 248--255. IEEE, 2009.

\bibitem{ganin2015unsupervised}
Yaroslav Ganin and Victor Lempitsky.
\newblock Unsupervised domain adaptation by backpropagation.
\newblock In {\em International Conference on Machine Learning}, pages
  1180--1189. PMLR, 2015.

\bibitem{gneiting2007strictly}
Tilmann Gneiting and Adrian~E Raftery.
\newblock Strictly proper scoring rules, prediction, and estimation.
\newblock {\em Journal of the American statistical Association},
  102(477):359--378, 2007.

\bibitem{gong2022robust}
Taesik Gong, Jongheon Jeong, Taewon Kim, Yewon Kim, Jinwoo Shin, and Sung-Ju
  Lee.
\newblock {NOTE}: Robust continual test-time adaptation against temporal
  correlation.
\newblock In Alice~H. Oh, Alekh Agarwal, Danielle Belgrave, and Kyunghyun Cho,
  editors, {\em Advances in Neural Information Processing Systems}, 2022.

\bibitem{grandvalet2004semi}
Yves Grandvalet and Yoshua Bengio.
\newblock Semi-supervised learning by entropy minimization.
\newblock {\em Advances in neural information processing systems}, 17, 2004.

\bibitem{guo2017calibration}
Chuan Guo, Geoff Pleiss, Yu Sun, and Kilian~Q Weinberger.
\newblock On calibration of modern neural networks.
\newblock In {\em International Conference on Machine Learning}, pages
  1321--1330. PMLR, 2017.

\bibitem{hendrycks2018benchmarking}
Dan Hendrycks and Thomas Dietterich.
\newblock Benchmarking neural network robustness to common corruptions and
  perturbations.
\newblock In {\em International Conference on Learning Representations}, 2019.

\bibitem{hoffman2018cycada}
Judy Hoffman, Eric Tzeng, Taesung Park, Jun-Yan Zhu, Phillip Isola, Kate
  Saenko, Alexei Efros, and Trevor Darrell.
\newblock Cycada: Cycle-consistent adversarial domain adaptation.
\newblock In {\em International Conference on Machine Learning}, pages
  1989--1998. Pmlr, 2018.

\bibitem{izmailov2018averaging}
Pavel Izmailov, Dmitrii Podoprikhin, Timur Garipov, Dmitry Vetrov, and
  Andrew~Gordon Wilson.
\newblock Averaging weights leads to wider optima and better generalization.
\newblock In {\em 34th Conference on Uncertainty in Artificial Intelligence
  2018, UAI 2018}, pages 876--885. Association For Uncertainty in Artificial
  Intelligence (AUAI), 2018.

\bibitem{kar20223d}
O{\u{g}}uzhan~Fatih Kar, Teresa Yeo, Andrei Atanov, and Amir Zamir.
\newblock 3d common corruptions and data augmentation.
\newblock In {\em Proceedings of the IEEE/CVF Conference on Computer Vision and
  Pattern Recognition}, pages 18963--18974, 2022.

\bibitem{kirkpatrick2017overcoming}
James Kirkpatrick, Razvan Pascanu, Neil Rabinowitz, Joel Veness, Guillaume
  Desjardins, Andrei~A Rusu, Kieran Milan, John Quan, Tiago Ramalho, Agnieszka
  Grabska-Barwinska, et~al.
\newblock Overcoming catastrophic forgetting in neural networks.
\newblock {\em Proceedings of the national academy of sciences},
  114(13):3521--3526, 2017.

\bibitem{koh2021wilds}
Pang~Wei Koh, Shiori Sagawa, Henrik Marklund, Sang~Michael Xie, Marvin Zhang,
  Akshay Balsubramani, Weihua Hu, Michihiro Yasunaga, Richard~Lanas Phillips,
  Irena Gao, et~al.
\newblock Wilds: A benchmark of in-the-wild distribution shifts.
\newblock In {\em International Conference on Machine Learning}, pages
  5637--5664. PMLR, 2021.

\bibitem{Krizhevsky09learningmultiple}
Alex Krizhevsky.
\newblock Learning multiple layers of features from tiny images.
\newblock Technical report, University of Toronto, 2009.

\bibitem{kundu2020universal}
Jogendra~Nath Kundu, Naveen Venkat, R~Venkatesh Babu, et~al.
\newblock Universal source-free domain adaptation.
\newblock In {\em Proceedings of the IEEE/CVF Conference on Computer Vision and
  Pattern Recognition}, pages 4544--4553, 2020.

\bibitem{lee2013pseudo}
Dong-Hyun Lee et~al.
\newblock Pseudo-label: The simple and efficient semi-supervised learning
  method for deep neural networks.
\newblock In {\em Workshop on challenges in representation learning, ICML},
  volume~3, page 896, 2013.

\bibitem{li2020model}
Rui Li, Qianfen Jiao, Wenming Cao, Hau-San Wong, and Si Wu.
\newblock Model adaptation: Unsupervised domain adaptation without source data.
\newblock In {\em Proceedings of the IEEE/CVF Conference on Computer Vision and
  Pattern Recognition}, pages 9641--9650, 2020.

\bibitem{li2016revisiting}
Yanghao Li, Naiyan Wang, Jianping Shi, Jiaying Liu, and Xiaodi Hou.
\newblock Revisiting batch normalization for practical domain adaptation, 2017.

\bibitem{li2017learning}
Zhizhong Li and Derek Hoiem.
\newblock Learning without forgetting.
\newblock {\em IEEE transactions on pattern analysis and machine intelligence},
  40(12):2935--2947, 2017.

\bibitem{liang2020we}
Jian Liang, Dapeng Hu, and Jiashi Feng.
\newblock Do we really need to access the source data? source hypothesis
  transfer for unsupervised domain adaptation.
\newblock In {\em International Conference on Machine Learning}, pages
  6028--6039. PMLR, 2020.

\bibitem{long2015learning}
Mingsheng Long, Yue Cao, Jianmin Wang, and Michael Jordan.
\newblock Learning transferable features with deep adaptation networks.
\newblock In {\em International Conference on Machine Learning}, pages 97--105.
  PMLR, 2015.

\bibitem{lopez2017gradient}
David Lopez-Paz and Marc'Aurelio Ranzato.
\newblock Gradient episodic memory for continual learning.
\newblock {\em Advances in neural information processing systems}, 30, 2017.

\bibitem{maddox2019simple}
Wesley~J Maddox, Pavel Izmailov, Timur Garipov, Dmitry~P Vetrov, and
  Andrew~Gordon Wilson.
\newblock A simple baseline for bayesian uncertainty in deep learning.
\newblock {\em Advances in Neural Information Processing Systems}, 32, 2019.

\bibitem{mccloskey1989catastrophic}
Michael McCloskey and Neal~J Cohen.
\newblock Catastrophic interference in connectionist networks: The sequential
  learning problem.
\newblock In {\em Psychology of learning and motivation}, volume~24, pages
  109--165. Elsevier, 1989.

\bibitem{mummadi2021test}
Chaithanya~Kumar Mummadi, Robin Hutmacher, Kilian Rambach, Evgeny Levinkov,
  Thomas Brox, and Jan~Hendrik Metzen.
\newblock Test-time adaptation to distribution shift by confidence maximization
  and input transformation.
\newblock {\em arXiv preprint arXiv:2106.14999}, 2021.

\bibitem{niu2022efficient}
Shuaicheng Niu, Jiaxiang Wu, Yifan Zhang, Yaofo Chen, Shijian Zheng, Peilin
  Zhao, and Mingkui Tan.
\newblock Efficient test-time model adaptation without forgetting.
\newblock In {\em International Conference on Machine Learning}. PMLR, 2022.

\bibitem{pan2010domain}
Sinno~Jialin Pan, Ivor~W Tsang, James~T Kwok, and Qiang Yang.
\newblock Domain adaptation via transfer component analysis.
\newblock {\em IEEE transactions on neural networks}, 22(2):199--210, 2010.

\bibitem{parisi2019continual}
German~I Parisi, Ronald Kemker, Jose~L Part, Christopher Kanan, and Stefan
  Wermter.
\newblock Continual lifelong learning with neural networks: A review.
\newblock {\em Neural Networks}, 113:54--71, 2019.

\bibitem{prabhu2021sentry}
Viraj Prabhu, Shivam Khare, Deeksha Kartik, and Judy Hoffman.
\newblock Sentry: Selective entropy optimization via committee consistency for
  unsupervised domain adaptation.
\newblock In {\em Proceedings of the IEEE/CVF International Conference on
  Computer Vision}, pages 8558--8567, 2021.

\bibitem{ratcliff1990connectionist}
Roger Ratcliff.
\newblock Connectionist models of recognition memory: constraints imposed by
  learning and forgetting functions.
\newblock {\em Psychological review}, 97(2):285, 1990.

\bibitem{schneider2020improving}
Steffen Schneider, Evgenia Rusak, Luisa Eck, Oliver Bringmann, Wieland Brendel,
  and Matthias Bethge.
\newblock Improving robustness against common corruptions by covariate shift
  adaptation.
\newblock {\em Advances in Neural Information Processing Systems},
  33:11539--11551, 2020.

\bibitem{sun2020test}
Yu Sun, Xiaolong Wang, Zhuang Liu, John Miller, Alexei Efros, and Moritz Hardt.
\newblock Test-time training with self-supervision for generalization under
  distribution shifts.
\newblock In {\em International Conference on Machine Learning}, pages
  9229--9248. PMLR, 2020.

\bibitem{taori2020measuring}
Rohan Taori, Achal Dave, Vaishaal Shankar, Nicholas Carlini, Benjamin Recht,
  and Ludwig Schmidt.
\newblock Measuring robustness to natural distribution shifts in image
  classification.
\newblock {\em Advances in Neural Information Processing Systems},
  33:18583--18599, 2020.

\bibitem{tsai2018learning}
Yi-Hsuan Tsai, Wei-Chih Hung, Samuel Schulter, Kihyuk Sohn, Ming-Hsuan Yang,
  and Manmohan Chandraker.
\newblock Learning to adapt structured output space for semantic segmentation.
\newblock In {\em Proceedings of the IEEE Conference on Computer Vision and
  Pattern Recognition}, pages 7472--7481, 2018.

\bibitem{wang2020tent}
Dequan Wang, Evan Shelhamer, Shaoteng Liu, Bruno Olshausen, and Trevor Darrell.
\newblock Tent: Fully test-time adaptation by entropy minimization.
\newblock In {\em ICLR}, 2021.

\bibitem{wang2021ordisco}
Liyuan Wang, Kuo Yang, Chongxuan Li, Lanqing Hong, Zhenguo Li, and Jun Zhu.
\newblock Ordisco: Effective and efficient usage of incremental unlabeled data
  for semi-supervised continual learning.
\newblock In {\em Proceedings of the IEEE/CVF Conference on Computer Vision and
  Pattern Recognition}, pages 5383--5392, 2021.

\bibitem{wang2018deep}
Mei Wang and Weihong Deng.
\newblock Deep visual domain adaptation: A survey.
\newblock {\em Neurocomputing}, 312:135--153, 2018.

\bibitem{wang2022continual}
Qin Wang, Olga Fink, Luc Van~Gool, and Dengxin Dai.
\newblock Continual test-time domain adaptation.
\newblock In {\em Proceedings of the IEEE/CVF Conference on Computer Vision and
  Pattern Recognition}, 2022.

\bibitem{xie2020self}
Qizhe Xie, Minh-Thang Luong, Eduard Hovy, and Quoc~V Le.
\newblock Self-training with noisy student improves imagenet classification.
\newblock In {\em Proceedings of the IEEE/CVF Conference on Computer Vision and
  Pattern Recognition}, pages 10687--10698, 2020.

\bibitem{xie2017aggregated}
Saining Xie, Ross Girshick, Piotr Doll{\'a}r, Zhuowen Tu, and Kaiming He.
\newblock Aggregated residual transformations for deep neural networks.
\newblock In {\em Proceedings of the IEEE Conference on Computer Vision and
  Pattern Recognition}, pages 1492--1500, 2017.

\bibitem{yang2020fda}
Yanchao Yang and Stefano Soatto.
\newblock Fda: Fourier domain adaptation for semantic segmentation.
\newblock In {\em Proceedings of the IEEE/CVF Conference on Computer Vision and
  Pattern Recognition}, pages 4085--4095, 2020.

\bibitem{zagoruyko2016wide}
Sergey Zagoruyko and Nikos Komodakis.
\newblock Wide residual networks.
\newblock In {\em British Machine Vision Conference 2016}. British Machine
  Vision Association, 2016.

\bibitem{zhou2021training}
Aurick Zhou and Sergey Levine.
\newblock Training on test data with bayesian adaptation for covariate shift.
\newblock {\em NeurIPS}, 2021.

\end{thebibliography}

\clearpage
\section*{\Large{Appendix}}
\appendix
In this Appendix, we provide further details about benchmark datasets, computational resources, training details, evaluation metrics and additional experimental analysis.
\section{Benchmark Datasets}
CIFAR10C and CIFAR100C datasets are corrupted versions of the standard CIFAR10 and CIFAR100~\cite{Krizhevsky09learningmultiple} datasets, respectively.
ImageNetC~\cite{hendrycks2018benchmarking} and ImageNet3DCC \cite{kar20223d} are both corrupted versions of the standard ImageNet~\cite{deng2009imagenet} dataset. 

Both CIFAR10C and CIFAR100C datasets contain 10,000 images per corruption type, making 150,000 images in total for each dataset. For ImageNetC dataset, there are 50,000 images for each corruption type.
CIFAR10C, CIFAR100C, and ImageNet-C datasets consists of 15 diverse corruption types with 4 extra corruption types for validation.
Every corruption has five different levels of severity.
The various corruptions, along with a brief description, are as follows: i. Gaussian noise: often occurs in low light; ii. Shot noise: electronic noise due to discrete nature of light; iii. Impulse noise: color equivalent of salt-and-pepper noise, and it may be due to bit errors; iv. Defocus blur: caused when a photograph is taken out of focus; v. Frosted Glass Blur: due to an image through a frosted glass window; vi. Motion blur: occurs when a camera is rapidly moving; vii.  Zoom blur: happens when a camera quickly approaches an object; viii.  Snow: visually an obscuring kind of precipitation; ix.  Frost arises when ice crystals adhere to windows; x. Fog:  Objects are cloaked in fog, which is rendered using the diamond-square algorithm; xi. Brightness: varying brightness with the sunshine intensity; xii. Contrast: depends on the lighting conditions and color of the photographed item; xiii. Elastic transformations: Small image areas are stretched or contracted via elastic transformations; xiv. Pixelation: happens when a low-resolution picture is upsampled; xv. JPEG: lossy image compression that results in compression artifacts.

Recently proposed by \cite{kar20223d}, the ImageNet 3D Common Corruptions (ImageNet3DCC) dataset exploits the geometry of the scene in transformations, resulting in more realistic corruptions. 
The Imagenet3DCC dataset has 50,000 images for each corruption type.
It consists of 12 different types of corruption, each with 5 severity levels. The corruptions are as follows: i. Near focus: changing the focus area to the near portion of the scene at random; ii. Far focus: randomly change the focus to the scene's far part; iii. Bit error: caused by an imperfect video transmission channel; iv. Color quantization: decreases the RGB image's bit depth; v. Flash: caused by positioning a light source near the camera; vi. Fog 3D: generated by using a standard optical model for fog; vii. H.265 ABR: codec H.265 for compression with Average Bit Rate control mode; viii. H.265 CRF: codec H.265 for compression with Constant Rate Factor (CRF) control mode; ix. ISO noise: noise using a Poisson-Gaussian distribution; x. Low light: simulate low-light imaging setting by lowering the pixel intensities and adding Poisson-Gaussian distributed noise; xi. XY-motion blur: the main camera is moving along the image XY-plane; xii. Z-motion blur: the main camera is moving along the image Z-axis.

These datasets are developed to serve as benchmarks for measuring robustness of classification models.

\begin{figure*}[!ht]
    \centering
    \includegraphics[scale=0.45]{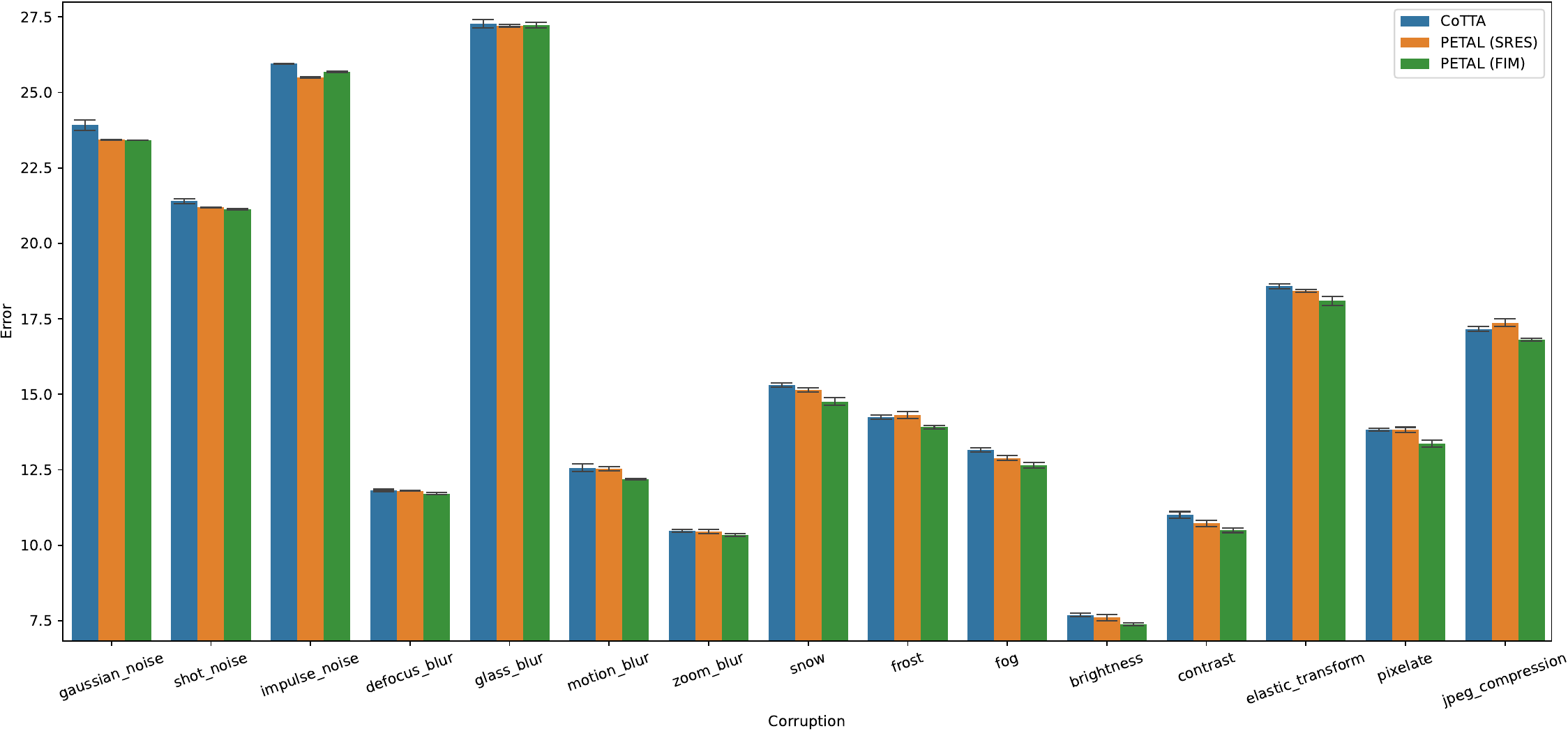}
    \caption{CIFAR10-to-CIFAR10C Results for the corruption order as depicted in the figure with the corruption of severity level 5. The error rate is averaged over 5 runs. The standard deviation is depicted in the bar plot. PETAL (FIM) performs better in most of the settings.}
    \label{fig:cifar10bar}
\end{figure*}
\begin{figure*}[!ht]
    \centering
    \includegraphics[scale=0.45]{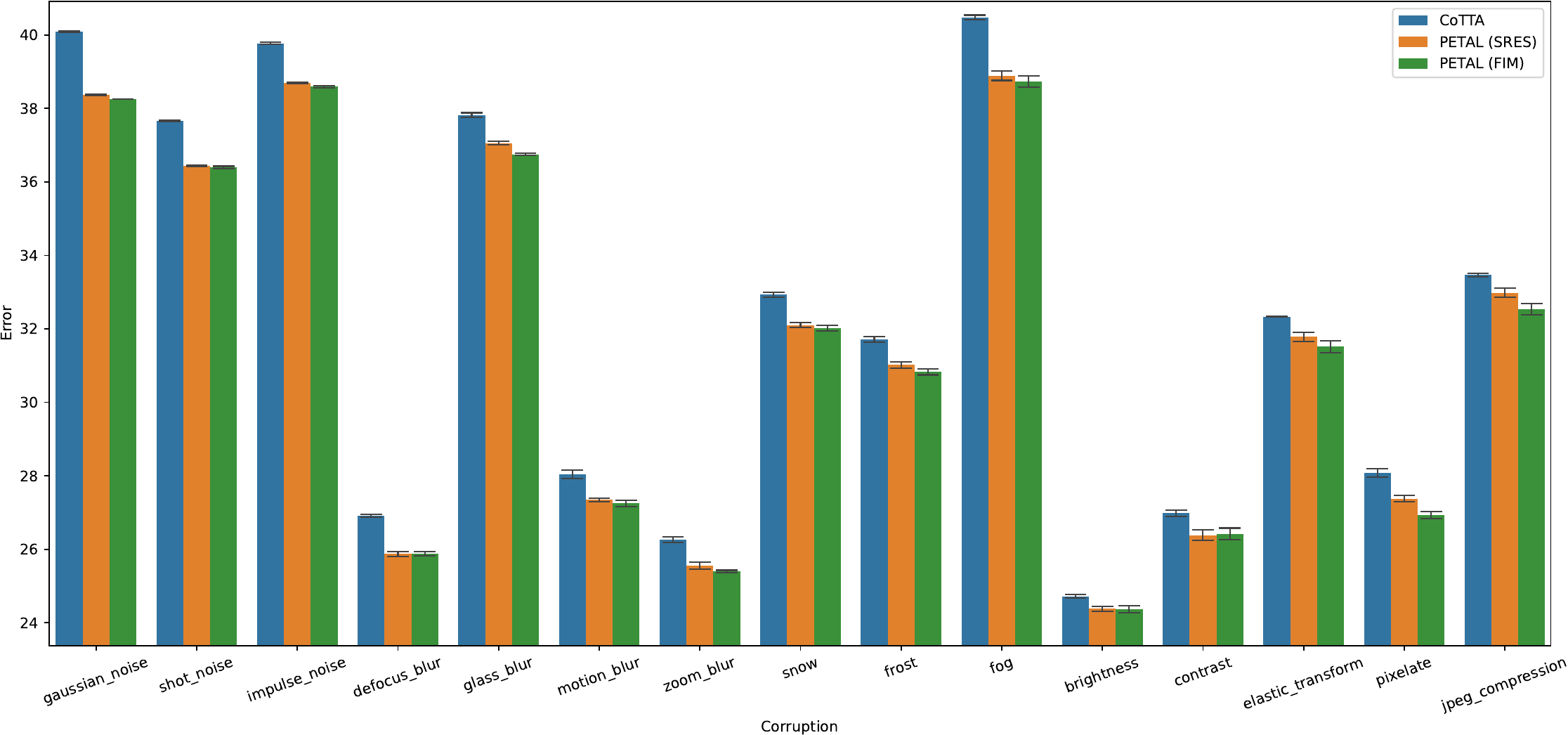}
    \caption{CIFAR100-to-CIFAR100C Results for the corruption order as depicted in the figure with the corruption of severity level 5. The error rate is averaged over 5 runs. The standard deviation is depicted in the bar plot. PETAL (FIM) performs better in most of the settings.}
    \label{fig:cifar100bar}
\end{figure*}
\section{Computational Resource Details}
All the experiments were run locally on a GPU server with Nvidia Titan RTX GPUs with 24 GB memory, Intel(R) Xeon(R) Silver 4110 CPU with 128 GB RAM, and Ubuntu 18.04.6 LTS OS. PETAL is implemented using PyTorch version 1.10.0. More details about the other relevant libraries are provided in the source code.

For CIFAR-10 experiments, further training the pre-trained WideResNet-28 model from RobustBench benchmark~\cite{croce2021robustbench} using the source domain training data takes approximately 20 minutes.
For CIFAR-100 experiments, we further train the pre-trained ResNeXt-29~\cite{xie2017aggregated} model from RobustBench benchmark~\cite{croce2021robustbench} using the source domain training data for approximately 15 minutes.
For ImageNet experiments, training the pre-trained ResNet50 model from RobustBench benchmark~\cite{croce2021robustbench} further using the source domain training data takes approximately 5 hours. 

For the CIFAR10-to-CIFAR10C adaptation experiment on the highest severity level of 5 using our approach PETAL, the total time taken for adapting the pre-trained WideResNet-28 model on 15 corruption types takes approximately 3 hours.
In CIFAR10-to-CIFAR10C gradual experiment, since the severity level changes gradually before the change in corruption type, the total number of different corruption-severity level pairs is 131.
Further, the experiment is conducted for 10 random orders of corruption sequences and so the total time taken is about 6 days and 19 hours.

CIFAR100-to-CIFAR100C adaptation experiment using pre-trained ResNeXt-29 for the highest severity level of 5 takes approximately 1 hour and 10 minutes.

For ImageNet-to-ImageNetC adaptation experiment on the highest severity level of 5, the total time taken for adapting the pre-trained ResNet50 model on 15 corruption types repeated for 10 diverse corruption sequences is about 2 hours and 30 minutes in total. The ImageNet-to-ImageNet3DCC adaptation experiment on the highest severity level of 5 for 12 corruptions took a total time of about 2 hours and 15 minutes.

\begin{figure*}[!htbp]
    \centering
    \includegraphics[scale=0.55]{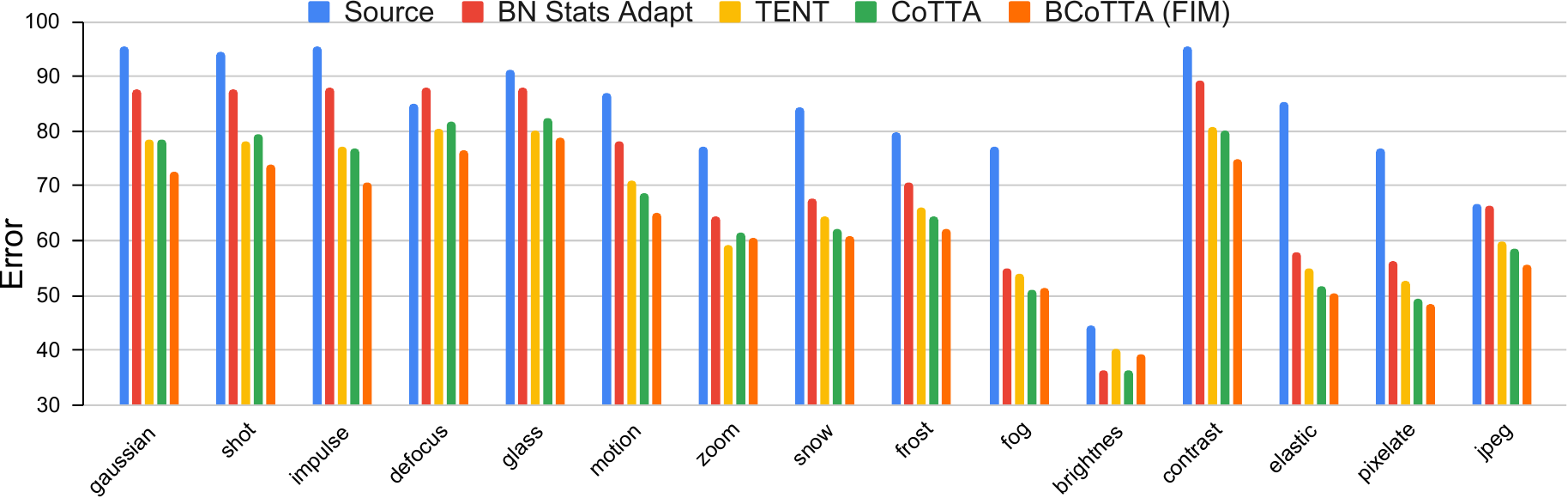}\\
    \vspace{0.5em}
    \includegraphics[scale=0.55]{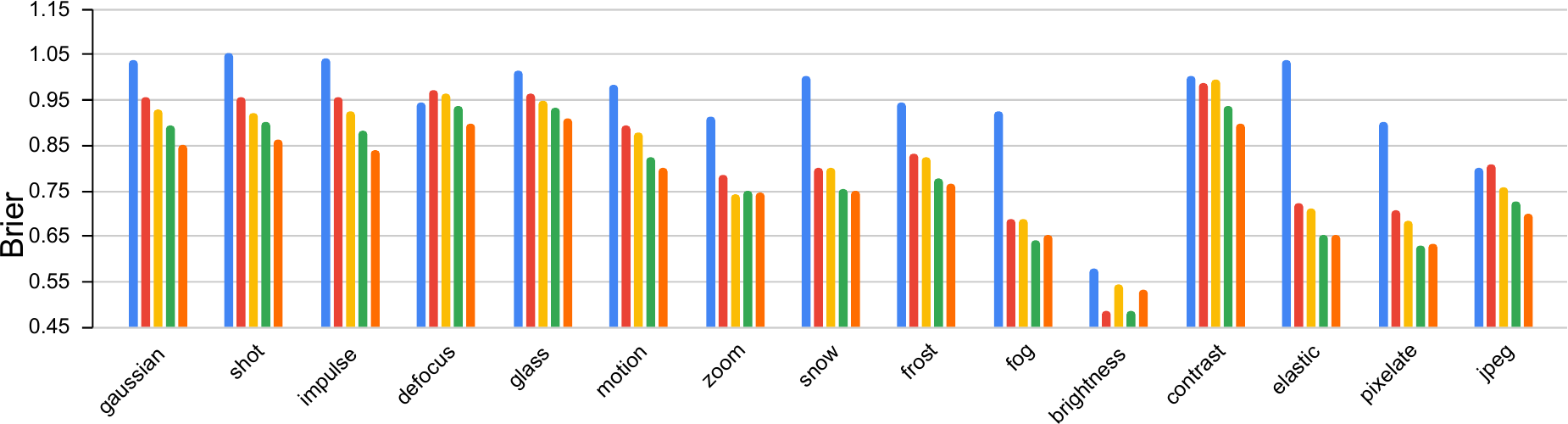}\\
    \vspace{0.5em}
    \includegraphics[scale=0.55]{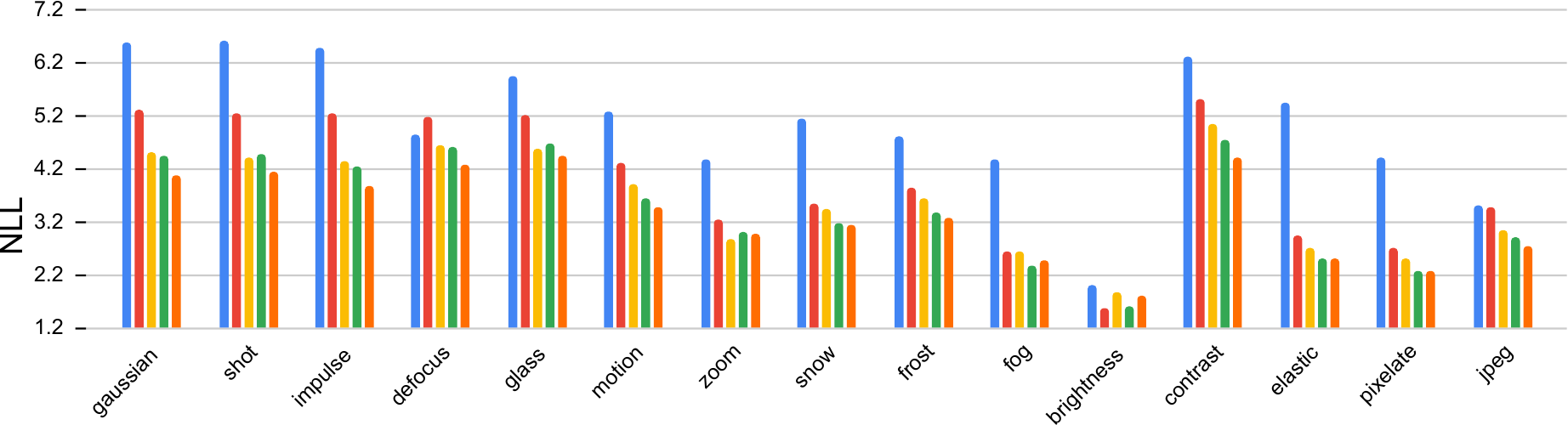}
    \caption{ImageNet-to-ImageNetC results averaged over 10 different corruption sequences with corruption of severity level 5. PETAL (FIM) outperforms other approaches for most of the corruptions.}
    \label{fig:imagenetbar}
\end{figure*}
\begin{figure*}[!htbp]
    \centering
    \includegraphics[scale=0.52]{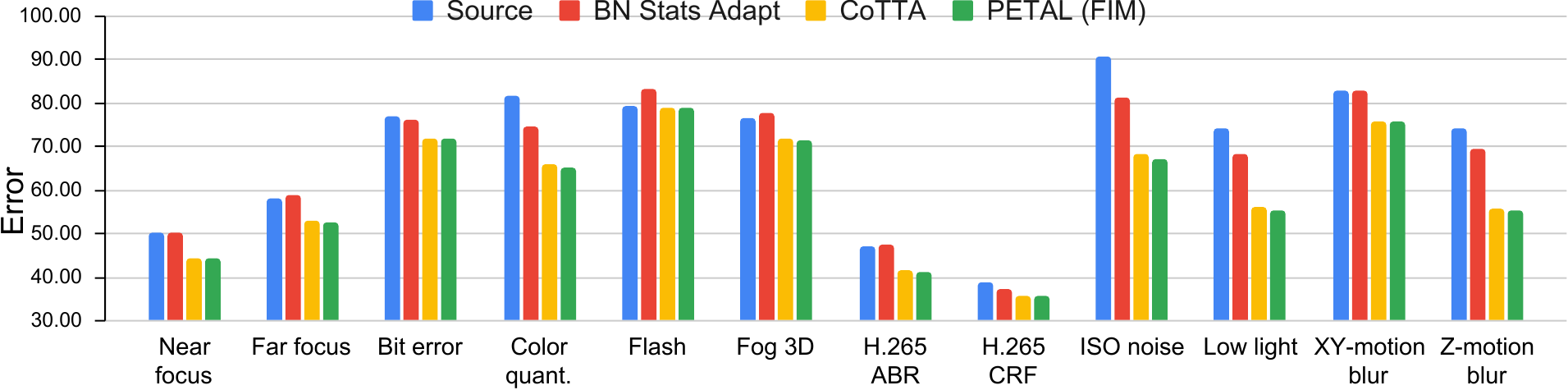}\\
    \vspace{0.5em}
    \includegraphics[scale=0.52]{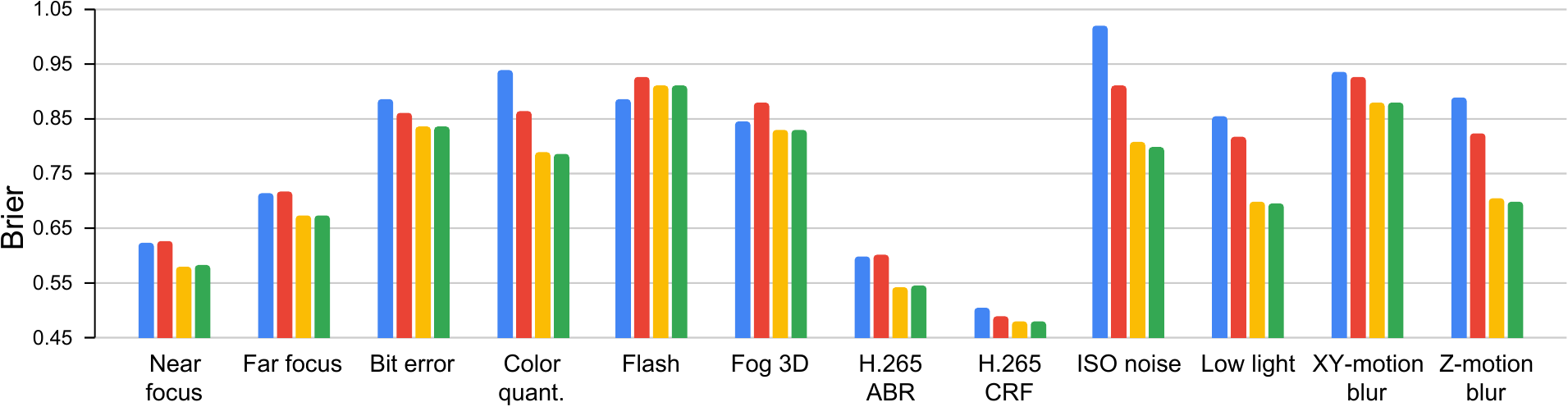}\\
    \vspace{0.5em}
    \includegraphics[scale=0.52]{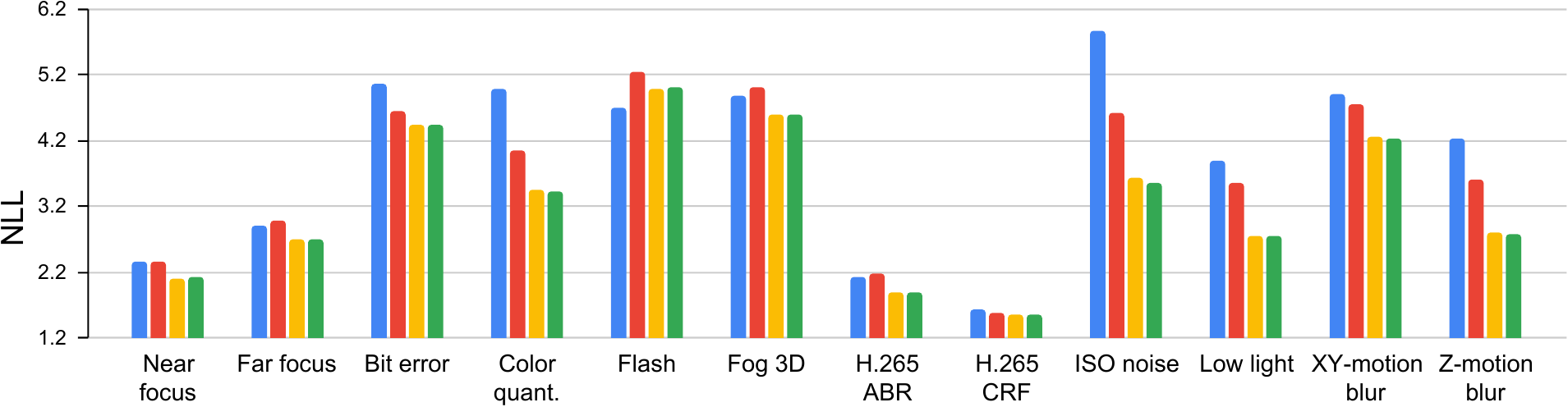}
    \caption{ImageNet-to-ImageNet3DCC results averaged over 10 different corruption sequences with corruption of severity level 5. PETAL (FIM) outperforms other approaches for most of the corruptions.}
    \label{fig:imagenet3dccbar}
\end{figure*}

\section{Training Details}
\subsection{Training on Source Domain data}
To obtain the approximate posterior using the source domain training data, we use SWAG-D~\cite{maddox2019simple} posterior.
This requires further training of the pre-trained models on the source domain training data for a few more epochs.

For CIFAR-10 experiments, the pre-trained WideResNet-28 model from RobustBench benchmark~\cite{croce2021robustbench} is used, which is already trained using CIFAR-10 training data.
We further train it using SGD with momentum for 5 epochs with a learning rate of 
8e-4, collecting iterates for SWAG-D~\cite{maddox2019simple} once in each epoch.
Similarly, for CIFAR-100 experiments, we use the pre-trained ResNeXt-29~\cite{xie2017aggregated} model from RobustBench benchmark~\cite{croce2021robustbench} by training it further using SGD with momentum for 5 more epochs with a learning rate of 8e-4. For ImageNet experiments, we further train the pre-trained standard ResNet50 model from RobustBench benchmark~\cite{croce2021robustbench} for 2 more epochs and collect 4 iterates per epoch with a learning rate of 1e-4 for obtaining SWAG-D approximate posterior.

\subsection{Adaptation on Target Domain Test data \label{targetadapt}}
For the online lifelong test-time adaptation (TTA), we use the same hyperparameters mentioned in CoTTA~\cite{wang2022continual} using a learning rate of 0.001 with Adam optimizer. Based on \cite{cohen2021katana} and ~\cite{wang2022continual}, we apply random augmentations that include color jitter, random affine, Gaussian blur, random horizontal flip, and Gaussian noise.
For stochastic restore probability, we use the same value used by CoTTA, i.e., 0.01 for CIFAR10-to-CIFAR10C/CIFAR100-to-CIFAR100C and 0.001 for ImageNet-to-ImageNetC experiments. We tune the FIM based parameter restoration quantile value $\delta$ using the extra four validation corruptions and use 0.03 for CIFAR10-to-CIFAR10C/CIFAR100-to-CIFAR100C, and use 0.003 for ImageNet-to-ImageNetC and ImageNet-to-ImageNet3DCC.

To adapt the model, we optimize the following training objective from Eq.~\ref{finobj},
\begin{equation}
\max_\theta \log q(\theta) -\frac{\bar{\lambda}}{M} \sum_{m=1}^M\Hce(y', y|\mathbf{\bar{x}}).
\end{equation}
Here, $q(\theta)$ is the approximate posterior density obtained using SWAG-D~\cite{maddox2019simple}. $\bar{\lambda}$ is a hyperparameter that determines the importance of the cross-entropy minimization term relative to the posterior term. For the implementation, putting $\alpha=1/\bar{\lambda}$, we rewrite the training objective as
\begin{equation}
\max_\theta \alpha \log q(\theta) -\frac{1}{M} \sum_{m=1}^M\Hce(y', y|\mathbf{\bar{x}}).
\end{equation}
For all the experiments, we tune the $\alpha$ hyperparameter using the corresponding extra four validation corruptions and use one of the values from \{1e-6, 1e-7, 1e-9, 1e-10, 5e-10, 1e-11, 5e-11, 1e-12\} for which the average error is the lowest. 
We follow CoTTA~\cite{wang2022continual} for setting the hyperparameters: augmentation threshold for confident predictions $\tau$ and exponential moving average smoothing factor $\pi$ = 0.999. CoTTA~\cite{wang2022continual} discusses choice of $\tau$ in detail in their supplementary. 
\vspace{0.5em}\\
\textbf{Gradually changing corruptions:} In this setting, the severity of corruption changes gradually before the change in corruption type. For example, if $A_s$, $B_s$, and $C_s$ are three different corruptions of severity level $s$, then there are 23 corruption-severity pairs in total. They arrive as follows: 

$A_5$$\rightarrow$$A_4$$\rightarrow$$A_3$$\rightarrow$$A_2$$\rightarrow$$A_1$$\xrightarrow[corruption]{change}$$B_1$$\rightarrow$$B_2$$\rightarrow$$B_3$$\rightarrow$$B_4$
$\rightarrow$$B_5$$\rightarrow$$B_4$$\rightarrow$$B_3$$\rightarrow$$B_2$$\rightarrow$$B_1$$\xrightarrow[corruption]{change}$$C_1$$\rightarrow$$C_2$$\rightarrow$$C_3$$\rightarrow$$C_4$
$\rightarrow$$C_5$$\rightarrow$$C_4$$\rightarrow$$C_3$$\rightarrow$$C_2$$\rightarrow$$C_1$

When the corruption type changes, the severity level is 1, and thus, the domain shift is gradual. In addition, distribution shifts within each corruption type are also gradual.

\section{Evaluation Metrics}
Let $y'_n$ be the predicted probability vector and $y_n$ be the true label ($y_{ni}=1$ if $i$ is the true class label, else $y_{ni}=0$) for input $x_n$. Let the number of examples be $N$ and the number of classes be $D$.
\subsection{Error}
The average error rate is defined as
\begin{equation}
    \text{Error} = \frac{1}{N}\sum_{n=1}^N \mathbb{I}(y'_n \neq y_n).
\end{equation}
where $\mathbb{I}()$ is the indicator function.
\subsection{Brier Score}
The average Brier score~\cite{brier1950verification} is defined as
\begin{equation}
    \text{Brier score} = \frac{1}{N}\sum_{n=1}^N\sum_{i=1}^D (y_{ni}'-y_{ni})^2.
\end{equation}
\subsection{Negative Log-Likelihood}
The average negative log-likelihood (NLL) is defined as
\begin{equation}
    \text{NLL} = -\frac{1}{N}\sum_{n=1}^N\sum_{i=1}^D (y_{ni}\log y_{ni}').
\end{equation}

\section{Additional Experimental Results}
\subsection{CIFAR10-to-CIFAR10C Results}
Figure~\ref{fig:cifar10bar} shows the results of CIFAR10-to-CIFAR10C experiments in which the corruption types arrive in the depicted order. The standard deviations are depicted in the figure. PETAL performs better in most of the settings. Moreover, PETAL (FIM) performs better than PETAL (SRES) demonstrating the effectiveness of data-driven FIM based parameter resetting.
\subsection{CIFAR100-to-CIFAR100C Results}
The CIFAR100-to-CIFAR100C adaptation results are shown in Figure~\ref{fig:cifar100bar} in which the corruption types arrive in the depicted order. The standard deviations are depicted in the figure. We observe that PETAL outperforms most of the approaches.
\subsection{ImageNet-to-ImageNetC Results}
For ImageNet-to-ImageNetC experiments, we provide the error rate, Brier score, and NLL in Figure~\ref{fig:imagenetbar}.
The experiments are conducted for 10 diverse corruption sequences. The numbers in Figure~\ref{fig:imagenetbar} are obtained by averaging over the 10 different orders of corruption sequences.
PETAL (FIM) outperforms CoTTA and PETAL (SRES) for most of the corruptions in terms of average error rate, average Brier score, and average NLL.
\subsection{ImageNet-to-ImageNet3DCC Results}
We report the results for ImageNet-to-ImageNet3DCC experiments in Fig.~\ref{fig:imagenet3dccbar}. The numbers are averaged over 10 random corruption orders. Our approach \ourmodel{} performs better than other approaches for most of the corruption types.

Results demonstrate that our probabilistic approach, in which the source model's posterior leads to a regularizer and the data-driven reset helps make an informed choice of weights to reset/keep, significantly outperforms a number of baselines, including CoTTA, which is specifically designed for continual test time adaptation.
\subsection{Experiments with Similar Resetting Rate}
We compare the performance of PETAL (FIM) and PETAL (SRes) with a similar resetting rate of 0.03. For CIFAR10-to-CIFAR10C, the average error rate over 5 runs for PETAL (SRes) with 0.03 restoring rate is 16.51$\pm$0.02, whereas that of PETAL (FIM) is 15.95$\pm$0.04. This indicates that FIM enables more parameters to be restored than stochastic restore (SRes).
\subsection{Effectiveness of Various Components}
In Table 1 and Table 2, the better performance of PETAL (FIM) than PETAL (S-Res) highlights the contribution of FIM for improved performance over the stochastic restore. 
For CIFAR10-to-CIFAR10C, the average error rate for PETAL \textbf{with no parameter restore} over 5 runs is 16.37$\pm$0.03, which is worse than having a parameter restore.
In addition, the contribution of the regularizer term (caused by the source model's posterior) is clear from the better performance of PETAL over CoTTA \cite{wang2022continual}, given that CoTTA is a special case of PETAL with no regularizer term.
\subsection{Limitation and Future Scope}
A limitation of our proposed approach is that we use an approximate posterior obtained using SWAG-diagonal, and we can experiment with better approximations for the posterior in future work. Further, we can explore the problem setting of lifelong TTA where we have a small number of examples per batch.

\end{document}